\documentclass[lettersize,journal]{IEEEtran}
\usepackage{amsmath,amsfonts}
\usepackage{algorithmic}
\usepackage{array}
\usepackage[caption=false,font=normalsize,labelfont=sf,textfont=sf]{subfig}
\usepackage{textcomp}
\usepackage{stfloats}
\usepackage{url}
\usepackage{verbatim}
\usepackage{graphicx}
\usepackage{booktabs}
\usepackage{bbding}
\usepackage{epstopdf}
\usepackage[colorlinks,
linkcolor=black,
anchorcolor=black,
citecolor=black
]{hyperref}
\graphicspath{{figures/}}

\def\BibTeX{{\rm B\kern-.05em{\sc i\kern-.025em b}\kern-.08em
    T\kern-.1667em\lower.7ex\hbox{E}\kern-.125emX}}
\usepackage{balance}
\begin{document}
\title{MATCNN: Infrared and Visible Image Fusion Method Based on Multi-scale CNN with Attention Transformer}
\author{Jingjing Liu, Li Zhang, Xiaoyang Zeng \IEEEmembership{Senior,~IEEE}, Wanquan Liu \IEEEmembership{Senior,~IEEE}, Jianhua Zhang \IEEEmembership{Senior,~IEEE}

\thanks{This work was supported in part by the National Natural Science Foundation of China under Grant 62204044, and in part by theState Key Laboratory of Integrated Chips and Systems under Grant SKLICS-K202302, and in part by the Department of Science and Technology of Guangdong Province under Grant 2021CX02G450, and in part by the Special funds for promoting high-quality industrial development in Shanghai under Grant JJ-ZDHYLY-01-23-0004. (Corresponding author: Jianhua Zhang.)}
\thanks{J. Liu, L. Zhang and J. Zhang are with the Shanghai Key Laboratory of Chips and Systems for Intelligent Connected Vehicle, School of Microelectronics, Shanghai University, Shanghai 200444, China (email: $\{$jjliu, zhangli3849,jhzhang$\}$@shu.edu.cn).}
\thanks{J. Liu and X. Zeng are with the State Key Laboratory of Integrated Chips and Systems, Fudan University, Shanghai 201203, China.(email:xyzeng@fudan.edu.cn)}
\thanks{W. Liu is with the School of Intelligent Systems Engineering, Sun Yat-sen University, Shenzhen 510006, China (email: liuwq63@mail.sysu.edu.cn).}
}
\maketitle

\begin{abstract}
While attention-based approaches have shown considerable progress in enhancing image fusion and addressing the challenges posed by long-range feature dependencies, their efficacy in capturing local features is compromised by the lack of diverse receptive field extraction techniques.
To overcome the shortcomings of existing fusion methods in extracting multi-scale local features and preserving global features, this paper proposes a novel cross-modal image fusion approach based on a multi-scale convolutional neural network with attention Transformer (MATCNN).
MATCNN utilizes the multi-scale fusion module (MSFM) to extract local features at different scales and employs the global feature extraction module (GFEM) to extract global features.
Combining the two reduces the loss of detail features and improves the ability of global feature representation.
Simultaneously, an information mask is used to label pertinent details within the images, aiming to enhance the proportion of preserving significant information in infrared images and background textures in visible images in fused images. Subsequently, a novel optimization algorithm is developed, leveraging the mask to guide feature extraction through the integration of content, structural similarity index measurement, and global feature loss.
Quantitative and qualitative evaluations are conducted across various datasets, revealing that MATCNN effectively highlights infrared salient targets, preserves additional details in visible images, and achieves better fusion results for cross-modal images.
The code of MATCNN will be available at \href{https://github.com/zhang3849/MATCNN.git}{https://github.com/zhang3849/MATCNN.git}.
\end{abstract}

\begin{IEEEkeywords}
Image fusion, Multi-scale convolutional neural network with attention Transformer (MATCNN), Multi-scale fusion module (MSFM), Global feature extraction module (GFEM), Optimization algorithm
\end{IEEEkeywords}

\section{Introduction}
\IEEEPARstart{C}{ross}-modal sensors have the capability to capture a wide range of information from a scene through the acquisition of various physical signals or wavelength ranges, thereby providing a more extensive and comprehensive understanding.
Image fusion involves amalgamating original images captured within the same scene by different sensors into a single fused image through comprehensive analysis, selection, or enhancement \cite{zhou2023gan}.
The image after fusion processing describes the scene more comprehensively and reliably and fully uses the complementarity between different sensor imaging systems, which is helpful for image processing and visual perception, and is widely used in military \cite{zhang2021polarization}, medical \cite{liang2019mcfnet}, monitoring \cite{bao2019transmitter}, and other fields \cite{rao2023tgfuse}.
Specifically, visible and infrared image fusion is a common instance of cross-modal image fusion approaches.
Infrared images rely on the concept of radiation and provide valuable information under extreme conditions, such as object positioning, but often lack detailed environmental information, such as texture \cite{liu2021learning}.
Visible images based on the principle of spectral reflection contain valuable scene texture details that conform to human visual perception.
Still, they are prone to losing significant information under night conditions \cite{guo2023mdfn,li2022mafusion}.
By fusing infrared and visible images, it is feasible to acquire valuable features from diverse origins, eliminate redundant information, and create fusion images that highlight salient objects and exhibit precise texture details \cite{ma2019infrared}.
Various approaches have been presented in cross-modal image fusion, encompassing the conventional approaches, the deep learning-based approaches, and the attention Transformer based approaches \cite{liu2023sgfusion,li2022cgtf}.

Conventional fusion methods are mainly founded on relevant mathematical models to obtain information at different levels of the origin cross-modal images, and then appropriate fusion strategies are manually devised to integrate the information.
Commonly used techniques include multi-scale transformation \cite{wang2020multi,liu2014region,liu2017structure}, saliency \cite{liu2017infrared,ma2017infrared}, subspace \cite{cvejic2007region,mou2013image}, sparse representation \cite{maqsood2020multi,zhang2018sparse,liu2020entropy}, and so on.
Zhang \textit{et al.} \cite{zhang2016adaptive} presented a fusion approach that effectively enhances image contrast and details via the non-subsampled contour transform.
The results showed that this method has several benefits, including multi-scale, multi-direction, and translation variance, significantly enhancing the overall quality of generated images.
Due to the weaknesses of sparse representation based approaches in preserving details, Liu \textit{et al.} \cite{liu2016image} presented an approach relying on convolutional sparse representation, which can effectively solve this issue and be significantly superior to the analytical representation based fusion method.
Overall, although traditional methods have achieved certain fusion effects in simple scene applications, their ability to extract significant features is limited when there are significant differences in features between regions, resulting in blurred edges of the fused image, lack of texture details, and inability to adapt to complex fusion scenes \cite{li2017pixel}.

In recent years, with deep learning has demonstrated effective feature extraction and expression abilities in computer vision, numerous deep learning-based approaches have been put forth.
Among them, the convolutional neural network (CNN) based methods are the most extensively employed, which utilizes the excellent local feature extraction ability of convolutional networks to realize precise fusion \cite{singh2019multimodal}.
By designing convolutional neural network structures and loss functions, Ma \textit{et al.} \cite{ma2021stdfusionnet} achieved end-to-end feature extraction, feature fusion, and image reconstruction models, reducing spatial complexity and avoiding tedious manual design of fusion strategies.
Although CNN based methods effectively extract local features from images, they perform poorly in extracting global features.
To extract multi-scale features from images, researchers have proposed a feature pyramid network (FPN) with CNN as the backbone network, but it suffers from information loss caused by channel compression.
Meanwhile, due to the weak generalization ability and the difficulty in obtaining satisfactory fusion results based on supervised CNN, this has compelled researchers to explore unsupervised models.
As the most important unsupervised model in recent years, generative adversarial networks (GAN) are extensively employed in cross-modal image fusion, and have achieved significant results \cite{ma2020pan,zhang2021gan}.
The GAN-based method achieves unsupervised training through adversarial games between the discriminator and generator, with the encoder utilized to extract features from cross-modal images and the decoder utilized to fuse features fusion to generate fused images.
Through the adversarial game, the generated fused images are forced to maintain maximum similarity with all input images, thereby achieving better fusion results.
However, the instability of adversarial games significantly affects the training stability of GAN-based fusion method \cite{zhang2021image}.
Meanwhile, unsupervised training introduces many loss functions, leading to an increase in hyperparameters and affecting the convergence of the overall training objective \cite{zhou2023gan}.

Recently, attention-based approaches have made promising advancements in global feature extraction.
To overcome the shortcomings of existing CNN-based approaches in capturing global features and solve the long-range dependence of features in image fusion, researchers have introduced an attention mechanism into image fusion \cite{tang2023datfuse}.
Transformer, through its unique multi-head self-attention (MSA) computation method, calculates the correlations between each pixel or block and other pixels or blocks, providing a highly global perspective and alleviating the local limitations of CNN-based approaches.
Nevertheless, while attention-based approaches solve the long-range dependency of features, the capacity to extract local features is weakened due to the absence of feature extraction methods with different receptive fields.
This may potentially result in blurring and distortion of the fusion results, particularly in terms of details.

CNN-based approaches are useful for obtaining local characteristics, while transformer-based approaches are superior for extracting global features.
Researchers have started experimenting with combining CNN and transformer to extract information from images.
Liu \textit{et al.} \cite{liu2022mfst} proposed an end-to-end self-adaptive transformer fusion method, which extracts features of different scales through a four layer encoder, fuses multi-scale features through a transformer fusion block, and finally reconstructs the fused image through a nested connection decoder.
Although this method can effectively extract multimodal features of images through CNN and achieve satisfactory fusion results, it does not utilize the advantages of transformer blocks in extracting global features and only uses them as fusion modules.
Zhang \textit{et al.} \cite{zhang2023deep} proposed a novel dehazing method that utilizes CNN based module and Transformer based module in parallel to capture local and global features, respectively.
Although this method combines the advantages of CNN and Transformer to achieve enough results, it only fuses the dehazed results obtained by the two modules in a straightforward way.
Huang \textit{et al.} \cite{huang2023brighten} proposed a low light image enhancement method consisting of CNN based branch and transformer based branch. In addition, they introduced a feature interaction module to achieve the interaction between local and global features, rather than simply fusing the results of the two branches.

\begin{figure*}[!t]
\centering
\includegraphics[width=6in]{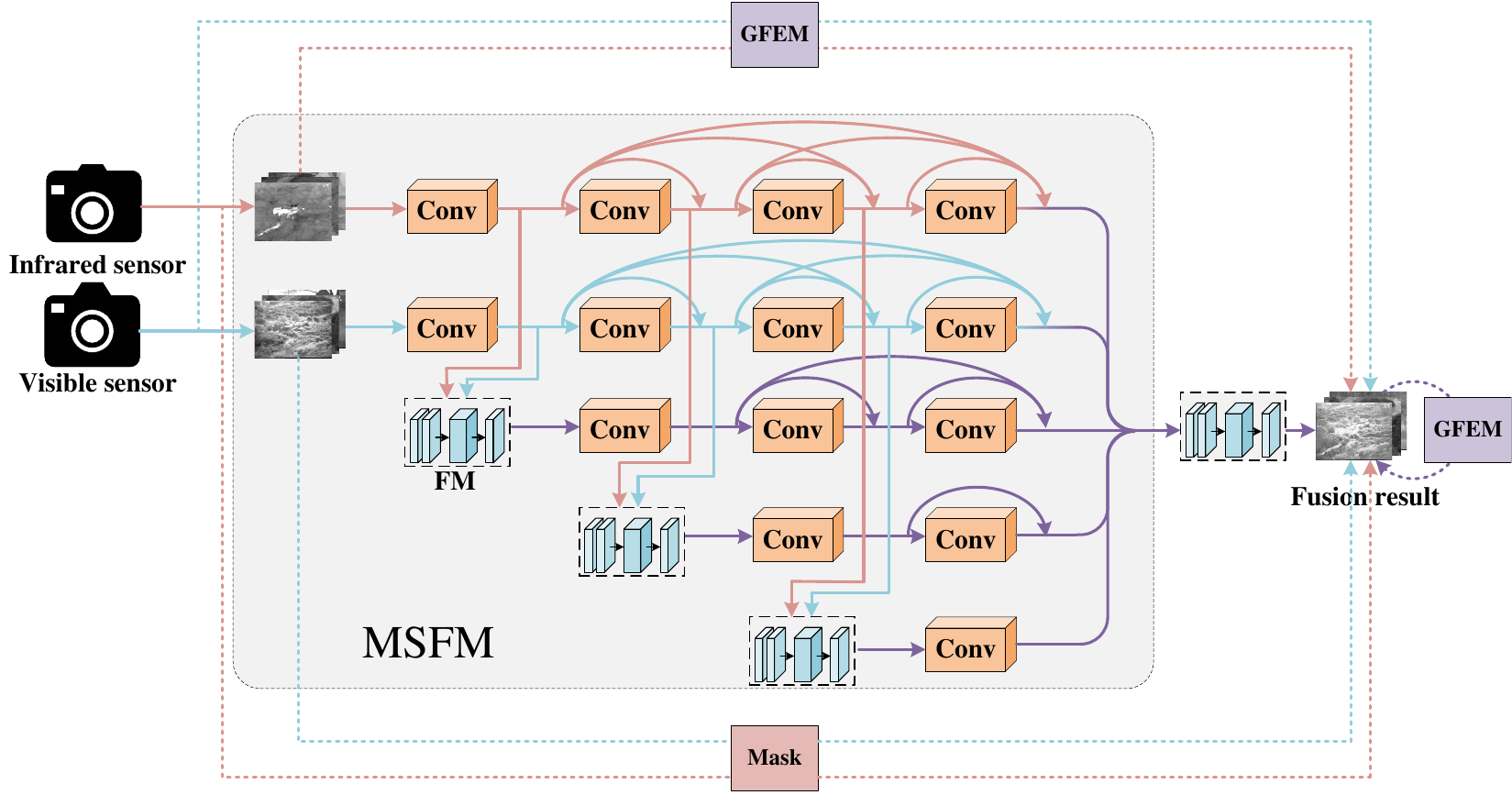}
\caption{Overview framework of the proposed MATCNN method.}
\label{fig_overview}
\end{figure*}

To address the aforementioned challenges, we propose a novel multi-scale convolutional neural network with an attention Transformer method for cross-modal image fusion in this paper.
The overview framework is shown in Fig. \ref{fig_overview}, where MSFM represents a multi-scale fusion module utilized for extracting and fusing multi-scale features.
In contrast, GFEM represents a global feature extraction module utilized to preserve global features.
The detail of GFEM is described in Fig. \ref{fig_swin}.
Unlike other methods combining CNN and transformer, the proposed method utilizes GFEM to measure the global feature differences between the original and fused images.
It constructs a global loss based on this to reduce the loss of important global features in the MSFM fusion process.
In addition, MATCNN has designed a salient information mask specifically for infrared images to guarantee the consistency of salient details in the fused image with the original infrared image while ensuring that texture information aligns with the visible image.

The main innovations of this paper are summarized in the following aspects,
\begin{itemize}
\item[$\bullet$] Present a novel cross-modal image fusion method based on MATCNN, which utilizes the MSFM to extract local features at various scales and employs the GFEM to capture global features, addressing the limitations of multi-scale local feature extraction while preserving global feature integrity.
\item[$\bullet$] Introduce a novel optimization algorithm that effectively preserves the prominent details and background texture of the original images, leveraging the mask to guide feature extraction through the integration of content, structural similarity index measurement, and global feature loss.
\item[$\bullet$] Evaluate on prevailing infrared and visible image datasets and compare with the latest approaches, the experiments demonstrate that the fusion results obtained via MATCNN exhibit more suitable saliency, fidelity, and contrast.
\end{itemize}

The rest of this paper is organized as follows.
Section \ref{related} provides an overview of the related works relevant to the topic of this paper, including CNN and attention Transformer.
A detailed illustration of MATCNN is presented in Section \ref{proposed}.
Both qualitative and quantitative experiments and analysis are detailed in Section \ref{experiments}.
Section \ref{conclusion} summarizes the work of the entire paper and provides conclusions.

\section{Related Work}\label{related}
\noindent
\subsection{CNN-based approaches}
Fukushima textit{et al.} proposed the neocognitron motivated by the human visual system, which many consider as the precursor to convolutional neural networks.
LeCun \textit{et al.} \cite{lecun1989handwritten} designed LeNet-5 for handwritten digit recognition, which was the earliest framework of a convolutional neural network.
However, due to limited computing power, there was no substantial progress in the subsequent decade.
Alex \textit{et al.} \cite{krizhevsky2017imagenet} proposed AlexNet, which demonstrated significant advantages in image classification tasks, marking the resurgence of neural networks, setting the prototype for today's CNN structures.
CNNs can extract abundant hierarchical information and have been popularly applied in different fields, including semantic segmentation, action recognition, and object detection, achieving significant results.

Liu \textit{et al.} \cite{liu2017multi} foremost applied CNN in image fusion. Their method involved learning weight maps using CNN and fusing the original pixels based on image segmentation principles.
However, although this method achieves satisfactory results in multi-focal image fusion, simple segmentation is not realistic because of the different imaging modes of different sensors, so the effect on the fusion of infrared and visible images is not ideal.
To tackle this issue, Liu \textit{et al.} \cite{liu2018infrared} presented a deep learning-based technique that is specifically designed for fusing infrared and visible images, which straight inputs the original images into the network to acquire weight maps.
The formulation of fusion strategies is learned autonomously by the network, providing a high level of autonomy and adaptability.
CNN, with its excellent capability for local feature extraction, exhibits strong performance in image fusion.
However, as the receptive field increases, the fusion of high-level semantic features not only leads to the loss of features at different scales but also results in the loss of long-range dependencies.

\subsection{Transformer-based approaches}
Transformer \cite{vaswani2017attention} was first proposed by Google and applied to Natural Language Processing (NLP), which attracted widespread attention and achieved remarkable achievements in multiple tasks in the field.
With the superior performance of Transformer based on global self-attention, more researchers have attempted to apply it to computer vision.
Dosovitskiy \textit{et al.} \cite{dosovitskiy2020image} proposed and introduced the specific architecture of the ViT model and clarified the feasibility of applying Transformer to computer vision.
ViT divides the input image into a succession of image patches, converts them into sequential data, and then processes them by the Transformer.
When sufficient data is available for pretraining, ViT has achieved performance surpassing traditional CNN models, bringing new ideas and methods to computer vision.
Inspired by VIT, Liu \textit{et al.} \cite{liu2021swin} introduced a model called Swin Transformer, which incorporates a hierarchical window attention mechanism into the traditional Transformer model, reducing computational complexity.
It divides the image into a series of windows and processes them using window-level attention while handling the details within each window using image patch-level attention.

The advantage of extracting long-range dependencies has led to the application of Transformer to image fusion in recent years.
In order to address the limitation of traditional CNN in capturing long-range dependencies, Vibashan \textit{et al.} \cite{vs2022image} proposed the Image Fusion Transformer (IFT).
This model introduces Transformer modules in the fusion layer, which enhances the fusion of long-range information while preserving local features.
Existing Transformer-based methods mainly explore intra-domain interactions.
To address this limitation, Ma \textit{et al.} \cite{Ma2022SwinFusion} proposed a Swin Transformer model that incorporates self-attention-based intra-domain fusion units and cross-attention-based inter-domain fusion units.
This design enables the integration of cross-domain dependencies and global interactions.
While Transformer exhibits excellent global performance and avoids generating artifacts, its ability to extract local features is less strong than CNN.
It may struggle to capture subtle variations between local features, potentially leading to blurriness or distortion in the fusion results at the detail level.

\subsection{Integration of multiscale local and global features.}
Local features capture the characteristics of key pixels or image regions, aiding in capturing specific objects and details within an image. Global features reflect the overall properties of an image, helping to capture the holistic structure of the image. Combining local and global features in computer vision allows for a more comprehensive description of image features, reducing noise interference and enhancing the richness and robustness of features.

In recent years, there has been a proliferation of work combining local features with global features.
Li \textit{et al.} \cite{li2022cgtf} proposed an infrared and visible light fusion model that alternates between convolutional and transformer layers.
This model combined the local features from convolutional networks with the long-range dependency features from transformers, yielding satisfactory results.
However, the model solely employs two branches to extract local and global features from the infrared and visible light images separately, without considering the interaction between features across different images during the extraction process.
Huo \textit{et al.} \cite{huo2024hifuse} proposed a three-branch hierarchical multiscale feature fusion model for medical image classification.
This model integrates a global feature module and a local feature module into a parallel hierarchical structure, utilizing an adaptive hierarchical feature fusion module to fuse multilayered global features with local features.
In addition, Yang \textit{et al.} \cite{yang2024multi} proposed multiscale dual attention (MDA) model for infrared and visible image fusion.
The model architecture consists of down-sampling blocks, attention-based fusion blocks, up-sampling blocks and reconstruction block.
The down-sampling module decomposes the original image into different scales, where each scale utilizes dual attention-based fusion blocks to extract and fuse features independently.
However, down-sampling blocks might lead to a loss of high-frequency details and fine-grained information in the images, which can affect the quality of the final fused output.

\section{The Proposed Method}\label{proposed}
To address the insufficient local feature extraction in the Transformer-based and the limitation of global feature extraction in the CNN-based fusion approach, a novel cross-modal image fusion network named MATCNN is presented in this paper.
The architecture of MATCNN is illustrated in Fig \ref{fig_overview}, where the network is composed of two innovative components, the multi-scale fusion module (MSFM) and the global feature extraction module (GFEM), with the two modules connected through a fusion module (FM).
In order to capture various types of information in cross-modal images, the trunk of the MSFM employs a pseudosiamese network, which utilizes four convolution operations to extract multi-scale features from the input images.
Furthermore, the MSFM comprises three branches for extracting fusion features at different scales.
Subsequently, the fusion features obtained from both the trunk and branches at different scales are combined through the fusion layer to generate the fused image.
Specifically, the fusion module concatenates features from different sources along the channel dimension, and employs a $1\times1$ convolution layer to fuse the concatenated features of different scales.
The design of the GFEM is based on the attention Transformer \cite{liu2021swin}, which is utilized to separately extract long-range features from the input and fused images, and employs a loss function to assess the global activity levels of pixels at four distinct scales.
The detailed structure and parameters of the proposed MSFM and GFEM will be elaborated in the following sections.
\subsection{Network architecture}
\subsubsection{MSFM}
The structure of the MSFM is illustrated in Fig.\ref{fig_overview}.
The fusion network consists of two trunks and three branches, achieving the fusion of input images at different scales.
The structure of the trunks pertains to the siamese convolutional neural network (SCNN) without parameter sharing, which extracts and encodes the features of the cross-modal images separately, ensuring that the parameters are learned to prevent feature confusion.
Each trunk network contains four convolutional layers that enable the extraction of features at four different scales.
In addition, each convolutional layer employs Relu activation and batch normalization (BN) layers.
In the first convolution layer, the number of the input and output channels is $1\sim64$, the convolution kernel size is $5\times5$, $stride=1$, and $padding=2$.
At the same time, dense connections are employed between the layers of the generator, which regard the output of each previous convolution layer as the input of the subsequent, facilitating feature reuse and minimizing feature loss.
As a result, the fused image contains multilevel information.
The number of input and output channels of the following three convolutional layers is $64\sim128$, $192\sim256$, and $448\sim512$, respectively.
The size of the convolutional kernel is $3\times3$, with $stride=1$ and $padding=1$.
The stride of these convolutional layers is all $1$, and the corresponding zero-filling mode ensures that each layer's output feature map size matches up with the input image, avoiding the introduction of transposed convolution in image reconstruction, reducing network parameters and avoiding the possible information loss caused by the up-sampling operation, thus improving the resulting image's quality in the process.

The input of the branch comes from the output of each convolutional layer of the trunk. Specifically, the outputs of the first three convolutional layers of the trunk are inputted to the fusion layer, and the features that pass through the fusion layer are inputted to the three branches of different scales.
In addition, the fusion layer adopts Tanh activation.
The fused feature maps on the branches should continue convolution for further feature extraction, and dense connections also adopted between the convolutional layers, each convolutional layer using a kernel of $3\times3$ with $stride=1$, $padding=1$.
For the scale $1$ branch, the fused features convolute thrice, and the number of input and output channels are $1\sim64$, $64\sim128$ and $128\sim256$.
For the scale $2$ branch, the fused features convolute twice, and the number of input and output channels are $1\sim64$, $64\sim128$.
For the scale $3$ branch, the number of output channels of the fused feature maps is $1\sim64$.
The feature maps derived from the trunks and branches were finally concatenated in channel dimensions and then input into the fusion layer to obtain the fused images.

\subsubsection{GFEM}

\begin{figure*}[!t]
\centering
\includegraphics[width=6.0in]{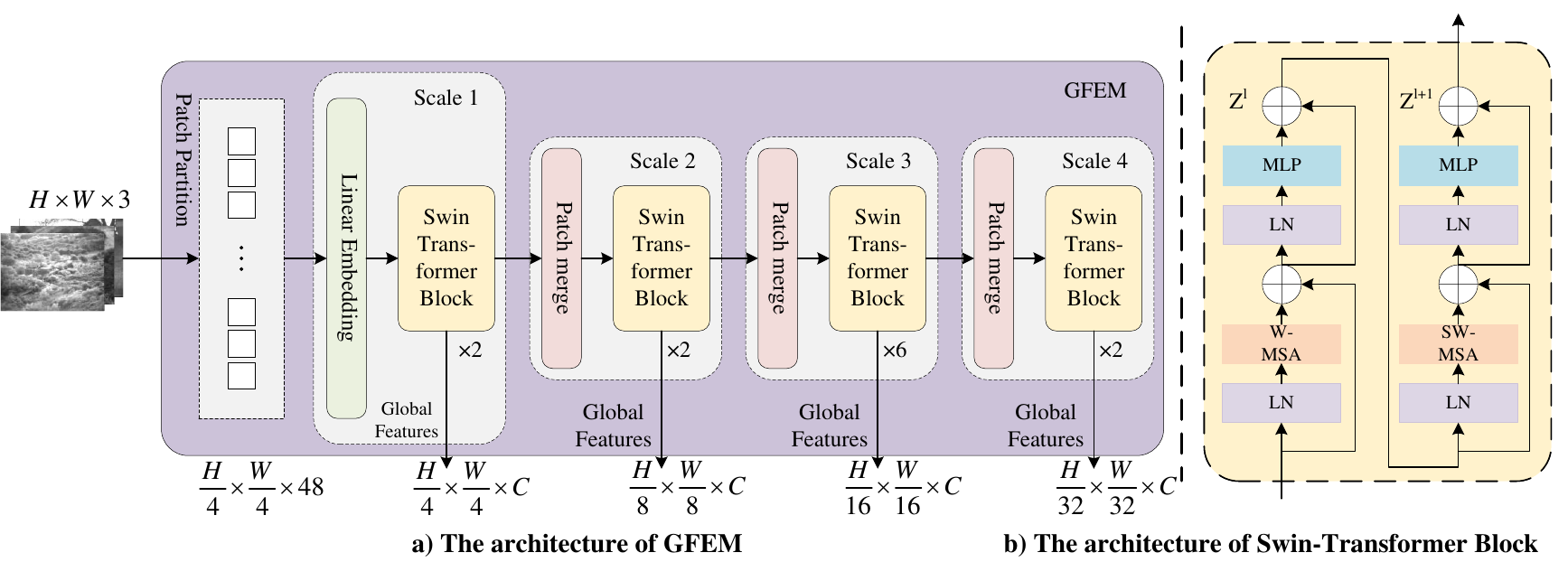}
\caption{The specific structure of the GFEM.}
\label{fig_swin}
\end{figure*}
After extracting the original image information through the MSFM and generating the fused image through the fusion layer, the GFEM module is used to extract the global features of the input and fused images at four scales, and then the L1 norm is utilized to measure the difference in global features between the input and fused images.
The GFEM is inspired by the Swin Transformer \cite{liu2021swin} and serves as the central component of the global feature extraction network and is depicted in Fig. \ref{fig_swin}.
The input single-channel image is first expanded into a three-channel image, then divided into $M \times M \times C$ local windows through the patch partition layer.
Subsequently, the window is linearly embedded into $HW/{M^2} \times M^2 \times C$ features through linear embedding, where $HW/{M^2}$ represents the total number of windows.
Then, the features are input into the transformer module to obtain the global features of the first scale.
In this paper, the training dataset is split into patches with $H = W = 128$, and $C = 3$, $M = 4$.
In order to obtain the global features of the second scale, the first scale global features are first input into the patch merging layer.
After fusion and convolution dimensionality reduction, they are input into the attention Transformer block of the second scale.
Finally, the second scale global features are obtained.
By performing the same operation on the third and fourth scales, global features can ultimately be obtained at four different scales.

The core of the GFEM is the transformer module, with the specific structure shown in the right box in Fig. \ref{fig_swin}.
Among the four scales, the third scale uses three pairs of attention Transformer structures, while the other three scales all use one pair of attention Transformer structures, and W-MSA and SW-MSA are used in each pair of structures respectively.
The encoded patch block is first utilized as input and preprocessed through the Layer Norm (LN) in layer $l$ for each attention Transformer structure.
Then, it is input into W-MSA to calculate the multi-head self-attention of each non-overlapping patch block.
Encode the input with the result of W-MSA through residual connection, and then input the result of residual connection into the LN, obtain the output through the Multi-Layer Perception (MLP) layer, and connect the output of the MLP layer with the previous residual connection again as the output of the $l$ layer, and then, input it to the $l+1$ layer.

\subsection{Loss function}
\begin{figure}
  \centering
  \includegraphics[width=3.1in]{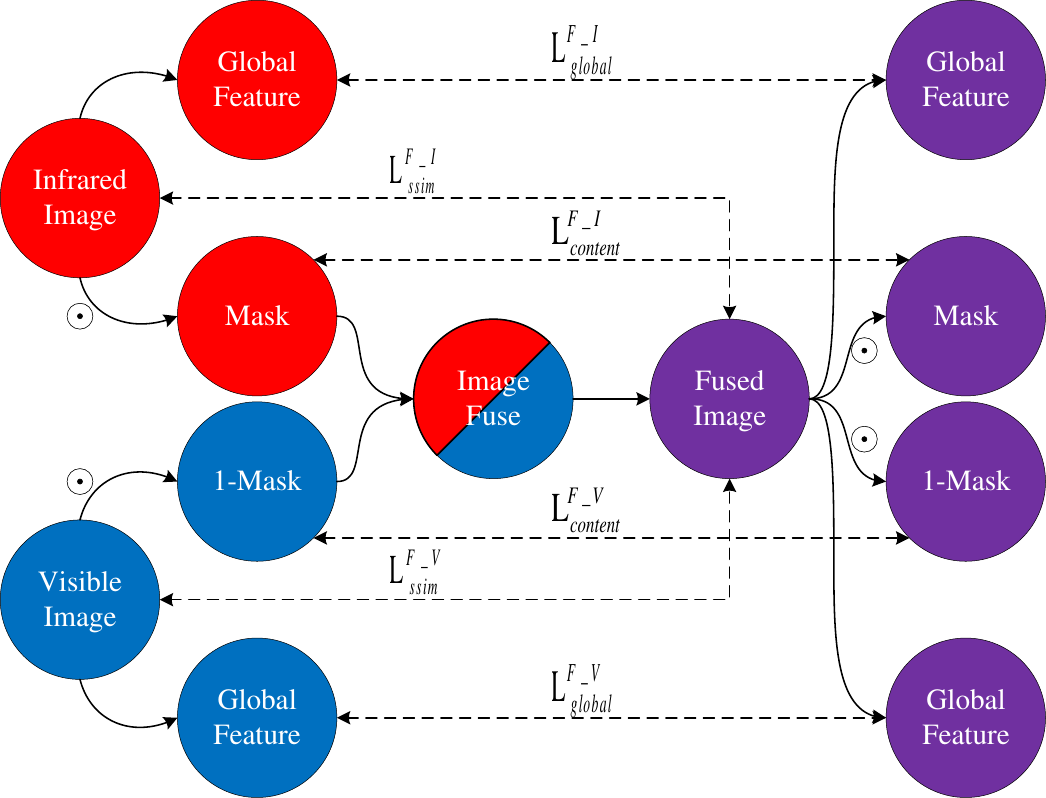}
  \caption{The overall loss function of MATCNN.}\label{ref_loss}
\end{figure}

This paper designs a novel loss function to maintain sufficient salient information and textural details in the generated image.
The overall loss function of MATCNN consists of three parts, which is apparent in Fig. \ref{ref_loss},
\begin{equation}\label{eq1}
\begin{aligned}
\boldsymbol{L}=\alpha \boldsymbol{L}_{content}+\beta\boldsymbol{L}_{ssim}+\gamma \boldsymbol{L}_{global},
\end{aligned}
\end{equation}
where $\boldsymbol{L}_{content}$ is referred to the content loss function,
$\boldsymbol{L}_{ssim}$ is referred to the structure similarity loss function, $\boldsymbol{L}_{global}$ is referred to the global loss function, and $\alpha$, $\beta$, $\gamma$ are referred to the weight hyper-parameters of each loss function.

Preserving crucial information from the infrared and visible images is paramount in the fusion process. Particularly, the salient objects in the infrared image and the background texture in the visible image are precious information sources. To address this, a content loss function is proposed in this paper, which integrates the salient object mask. This designed function effectively retains and incorporates valuable information from the infrared and visible images,

\begin{equation}\label{eq2}
\begin{aligned}
\boldsymbol{L}_{content}=\boldsymbol{L}_{content}^{F\_I}+\boldsymbol{L}_{content}^{F\_V},
\end{aligned}
\end{equation}
where $\boldsymbol{L}_{content}^{F\_I}$ is the content loss between the fused image and the infrared image, and $\boldsymbol{L}_{content}^{F\_V}$ is the content loss between the fused image and the visible image. More specific representations are as follows,

\begin{equation}\label{eq3}
\begin{aligned}
&\boldsymbol{L}_{content}^{F\_I}=\left\| I_m\circ(I_f-I_i) \right\| _{F}\\
&\boldsymbol{L}_{content}^{F\_V}=\left\| (1-I_m)\circ(\nabla I_f-\nabla I_v) \right\| _{F},
\end{aligned}
\end{equation}
where $I_f$ is the fused image, $I_i$ is the infrared image, $I_v$ is the visible image, $I_m$ is the salient object mask, $\circ$ denotes the elementwise multiplication operator, $\nabla$ denotes the gradient operator, and $\left\| \cdot \right\| _{F}$ calculates the matrix Frobenius norm.

SSIM (Structural Similarity Index) is an indicator that reflects the resemblance between the fused image and the original image in terms of brightness, contrast, and structure.
To ensure that the fused image maintains the structural properties of the cross-modal images while enhancing the general quality and fidelity of the fused image, MATCNN adopts SSIM loss,

\begin{equation}\label{eq4}
\begin{aligned}
\boldsymbol{L}_{ssim}=\boldsymbol{L}_{ssim}^{F\_I}+\boldsymbol{L}_{ssim}^{F\_V},
\end{aligned}
\end{equation}
where $\boldsymbol{L}_{ssim}^{F\_I}$ and $\boldsymbol{L}_{ssim}^{F\_V}$ represented the SSIM loss between fused images and infrared and visible images, respectively. The following are more detailed representations,

\begin{equation}\label{eq5}
\begin{aligned}
&\boldsymbol{L}_{ssim}^{F\_I}=1- SSIM(I_f,I_i) \\
&\boldsymbol{L}_{ssim}^{F\_V}=1- SSIM(I_f,I_v),
\end{aligned}
\end{equation}
where $\it{SSIM(\cdot)}$ denots the structural similarity operation.

To address the limitations associated with feature extraction and fusion tasks using CNN, a GFEM based on attention Transformer is proposed to extract global features of original and fused images at multiple scales.
To measure the pixel global activity level of the global features, the global feature loss $\boldsymbol{L}_{global}$ is designed, which can be expressed as,
\begin{equation}\label{eq9}
\begin{aligned}
\boldsymbol{L}_{global}=\sum_{i=1}^4{\left\| GF_i-\max \!\:\left( GA_i,GB_i \right) \right\| _1},
\end{aligned}
\end{equation}
where $GF$, $GA$, and $GB$ represent the 4-scale global features of the fused image and the input images, respectively.
In addition, $max(\cdot)$ denotes the maximum operator.
In the global feature loss, the larger value of the 4-scale global features of the input images is calculated, and the global activity level between it and that in the fused image is measured by the L1 norm to display more significant global features.
Global feature loss can measure the pixel global activity level by quantifying the global feature difference between the fused image and the input images.
This approach effectively enhances the feature continuity of the fused image and mitigates the loss of important global features during the fusion process.

Overall, the content loss function ensures the overall similarity between the fused image and the original image, the SSIM loss function ensures the structural similarity between the fused image and the original image, and the global loss function ensures that the fused image and the original image maintain the same global features.

\section{Experiments and Analysis}\label{experiments}
\subsection{Experiment and parameter settings}
Three datasets, which include the TNO, MSRS, and RoadScene datasets, are used to validate MATCNN.
The TNO dataset is a well-known and commonly utilized dataset in the area of cross-modal images fusion. It encompasses 60 pairs of cross-modal images that are specifically relevant to military scenarios. These images cover various environments, including indoor and outdoor, visible at night, challenging lighting conditions, adverse weather conditions, and other diverse situations.
The MSRS dataset is a cross-modal image dataset built on the MFNet.  It comprises 1444 pairs of precisely registered infrared and visible images, which is divided into a training set, consisting of 1083 pairs of images, and a test set, comprising 361 pairs of images.
The RoadScene dataset is a comprehensive cross-modal image dataset that includes 221 pairs of cross-modal images. This dataset exhibits precise registration and rich scene content, specifically focusing on elements associated with road environments, including roads themselves, vehicles, pedestrians, and other relevant objects.
In order to generate more samples for training, 46 pairs of strictly registered images from the TNO dataset are chosen for this paper.
The dataset was further processed by dividing the images into 41,703 pairs of patches, each with a size of 128$\times$128 pixels and a step size of 16 pixels.  Furthermore, multiple tests were conducted on the TNO dataset. For each test, 10 pairs of infrared and visible light images were randomly sampled using a systematic random sampling method.
To assess the performance of the MATCNN model on different datasets and demonstrate its generalization ability and accuracy, the model trained on the TNO dataset was tested on the MSRS and RoadScene datasets without the need for retraining.

To evaluate MATCNN comprehensively, comparing it qualitatively and quantitatively with the eight latest approaches, including GFF \cite{li2013image}, DenseFuse \cite{li2018densefuse}, FusionGAN \cite{ma2019fusiongan}, GANMcC \cite{ma2020ganmcc}, SDNet \cite{zhang2021sdnet}, SwinFusion \cite{Ma2022SwinFusion}, SeAFusion \cite{tang2022image}, PIAFusion \cite{Tang2022PIAFusion} and TUFusion \cite{zhao2023tufusion}.
The performance evaluation of fused images can be categorized into qualitative evaluation and quantitative evaluation.
Qualitative evaluation relies on human vision and can intuitively reflect aspects of fused images such as contrast, but more subjective.
Quantitative evaluation characterizes images through various evaluation indicators, which can evaluate image quality from different perspectives. In this paper, the following widely used indicators for evaluation are selected,

\textbf{EN:} entropy, aims to quantify the information content and visual quality of fused images. The higher value indicates that the fusion approach has a stronger information retention capacity and can better maintain the structures and features in the original images.

\textbf{SD:} standard deviation, quantifies the extent of change in the gray-scale or color distribution of the fused images. A higher standard deviation suggests that the fused images have more detail features and changes.

\textbf{SF:} space frequency, characterizes the details and texture and measures the gradient of the fused images. The fused images are sharper and have richer features and edge texture at higher spatial frequencies.

\textbf{VIF:} visual information fidelity, employs natural scene statistical models and human visual systems to assesses the quality of fused images. The higher the fidelity of visual information, the more suitable the fused images are for human visual perception.

\pmb{$Q^{AB/F}$:} edge retention, assesses the overall quality of the fusion approach via measuring the edge information that is transmitted from the original to the fused images.
A higher value indicates greater preservation of detailed and accurate edge structures from the original images in the fused image, leading to a higher-quality fusion outcome.

\textbf{MI:} mutual information, quantifies the total quantity of information that is communicated from the original images based on information theory metrics.
A higher MI score indicates greater preservation of relevant details and features from the original images in the fused image.

\begin{figure}
  \centering
  \includegraphics[width=3.5in]{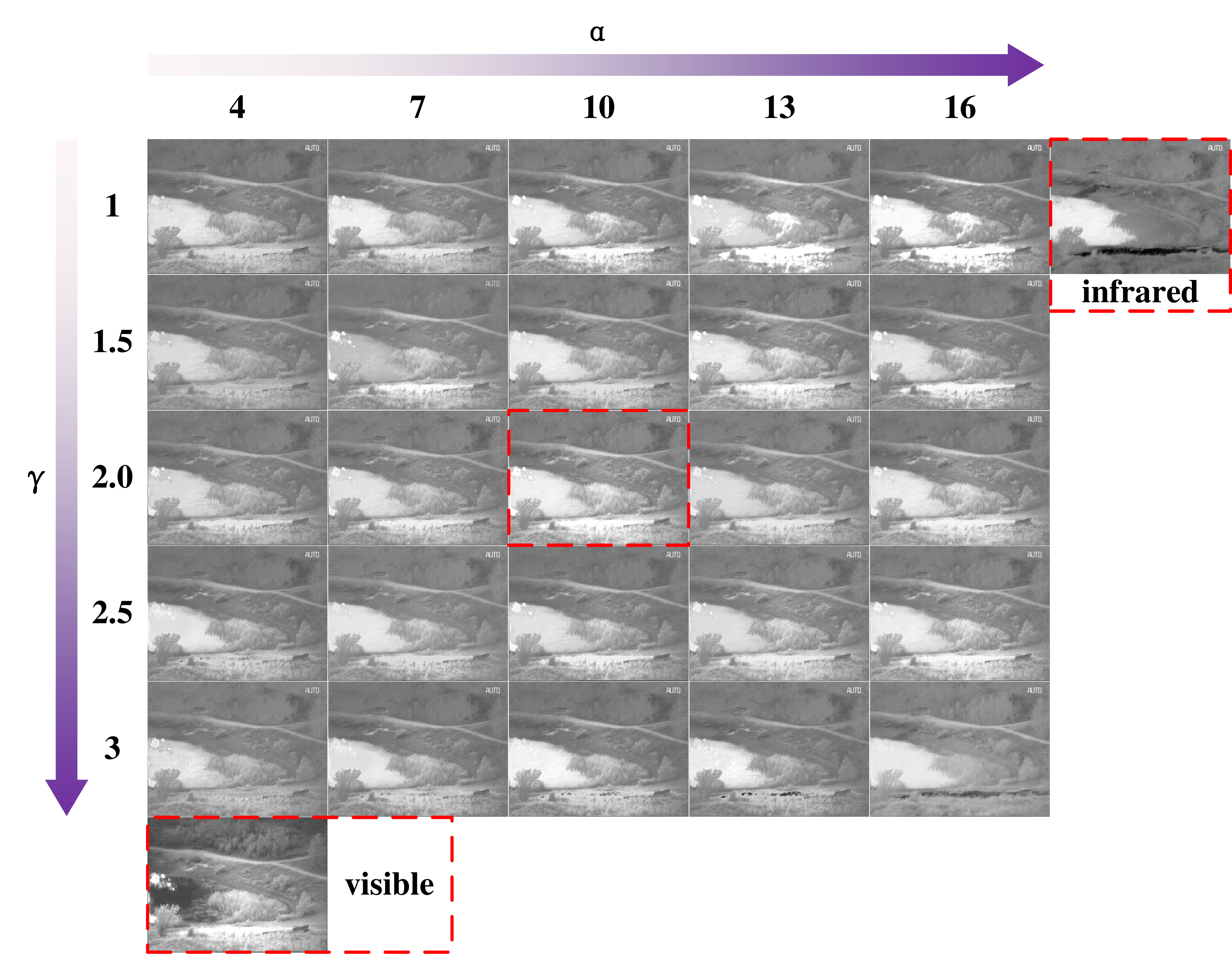}
  \caption{Fusion results with different $\alpha$ and $\gamma$.}\label{parameter}
\end{figure}

In the proposed loss function, the magnitude of the content loss function is smaller than that of the structure similarity loss function and global loss functions.
To balance the order of magnitude of each loss function, set the hyper-parameters $\alpha = 10$, $\beta = 1$, $\gamma = 1$ in the initial analysis.
Subsequently, analyzing the influence of hyper-parameters $\alpha$ and $\gamma$ values on the fusion results under the condition that the value of $\beta$ is fixed to 1, and the training iteration epoch is set to 10.
Their values are set as follows: $\alpha \in ({4, 7, 10, 13, 16})$, $\gamma \in ({1.0, 1.5, 2.0, 2.5, 3.0})$.
The outcome of different $\alpha$ and $\gamma$ values is displayed in Fig. \ref{parameter}, when $\alpha = 10$, $\gamma = 2$, the fused image is most in line with human visual habits.

In the training phase, the Adam optimizer is utilized by MATCNN.
The batch size is set to 32, and the total training iteration epoch is set to 110.
The training rate $lr$ is set to 0.2 in the first 100 iterations and then linearly decreases to 0.05 in the final 10 iterations.
As previously discussed,setting the hyper-parameters $\alpha = 10$, $\beta = 1$, $\gamma = 2$.
Furthermore, the PyTorch framework is used to train and test the approach on an Ubuntu operating system server that has an Intel Core i7-3770k CPU and an NVIDIA GeForce RTX 4090 GPU.

\subsection{Experiments on the TNO dataset}
\subsubsection{Qualitative evaluation}
\begin{figure}
  \centering
  \includegraphics[width=3.5in]{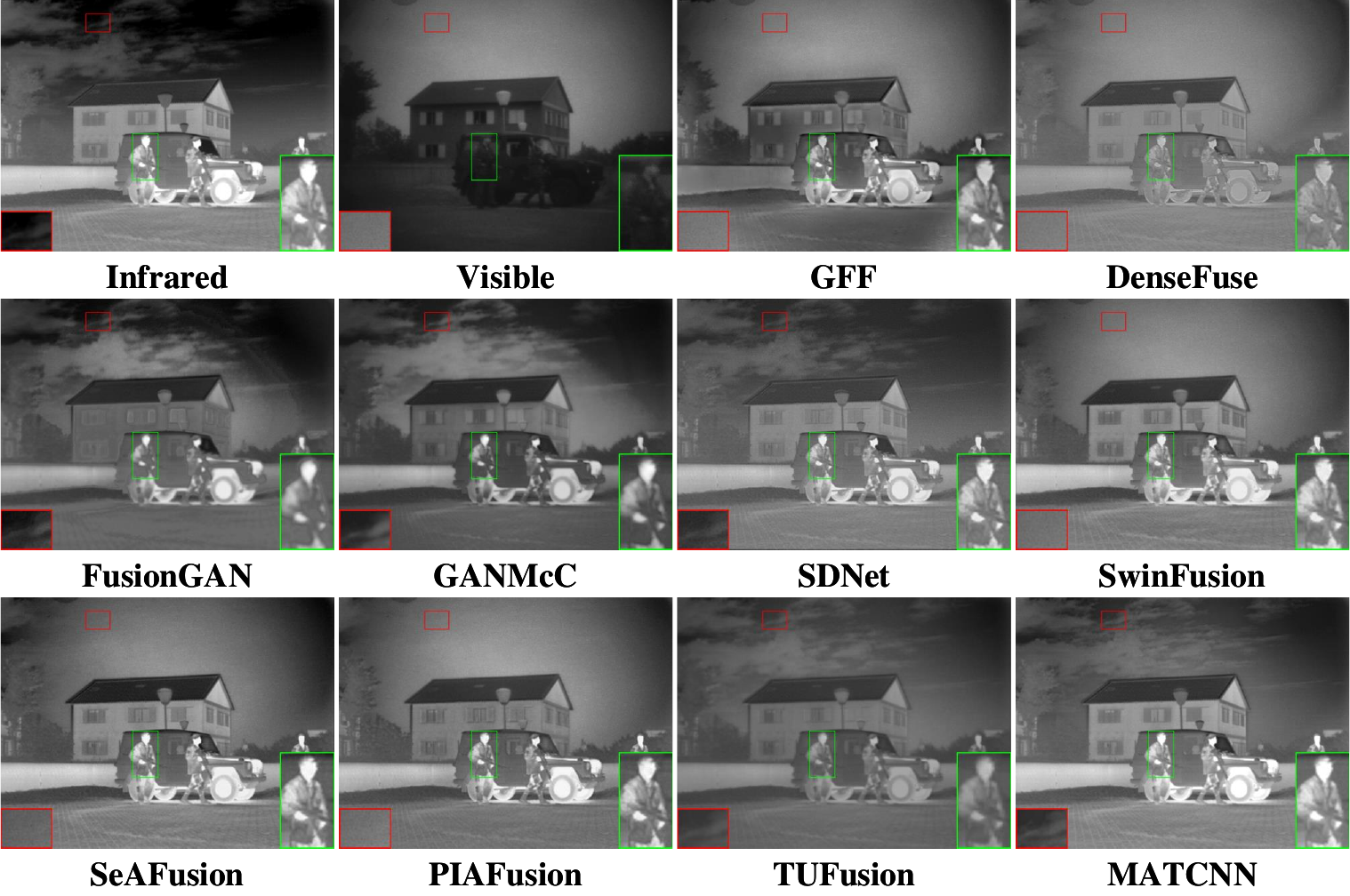}
  \caption{Fusion results of soldiers$\_$with$\_$jeep in the TNO dataset.}\label{ref_jeep}
\end{figure}

\begin{figure}
  \centering
  \includegraphics[width=3.5in]{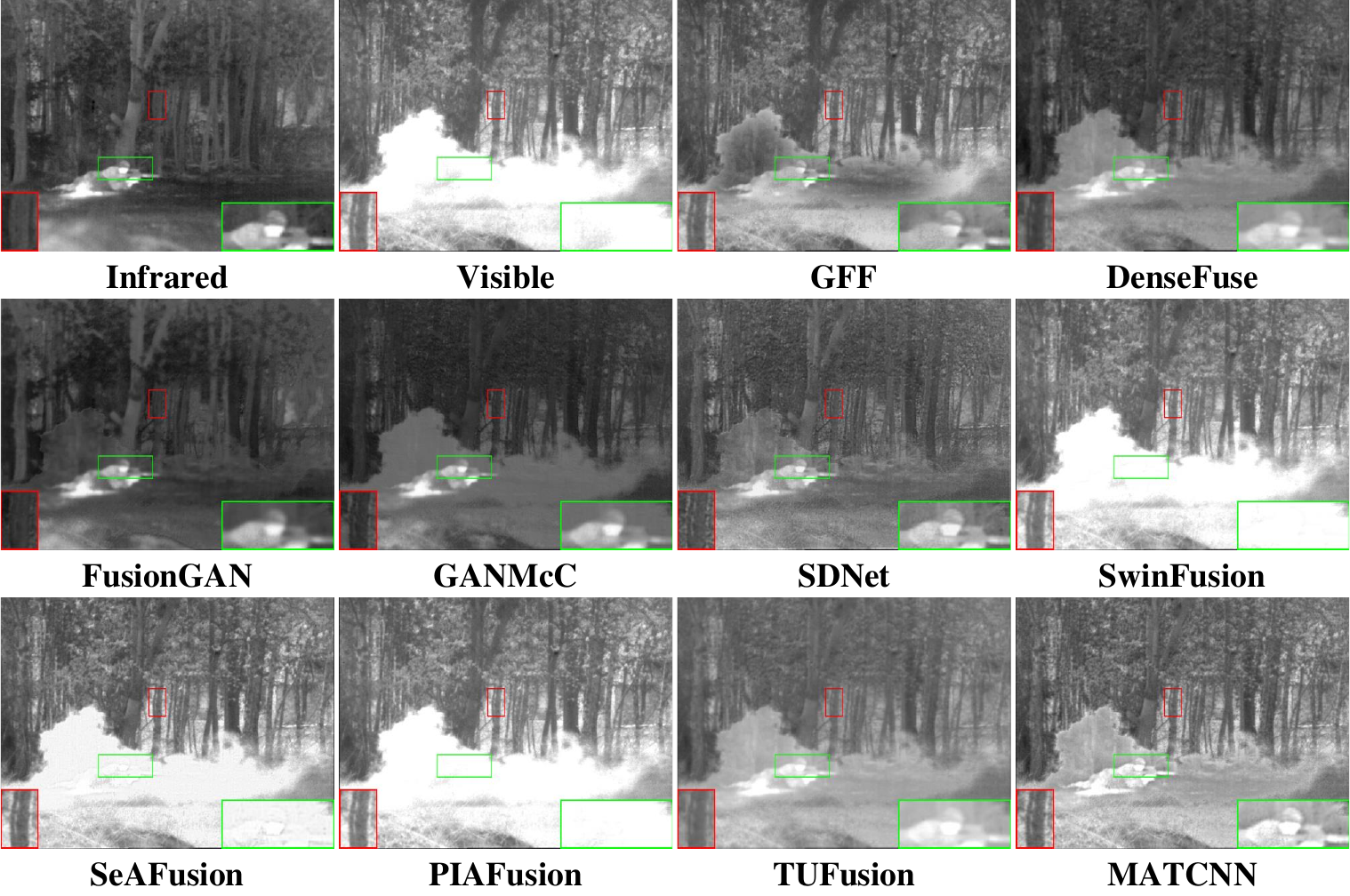}
  \caption{Fusion results of soldier$\_$behind$\_$smoke1 in the TNO dataset.}\label{ref_soldier}
\end{figure}

\begin{figure}
  \centering
  \includegraphics[width=3.5in]{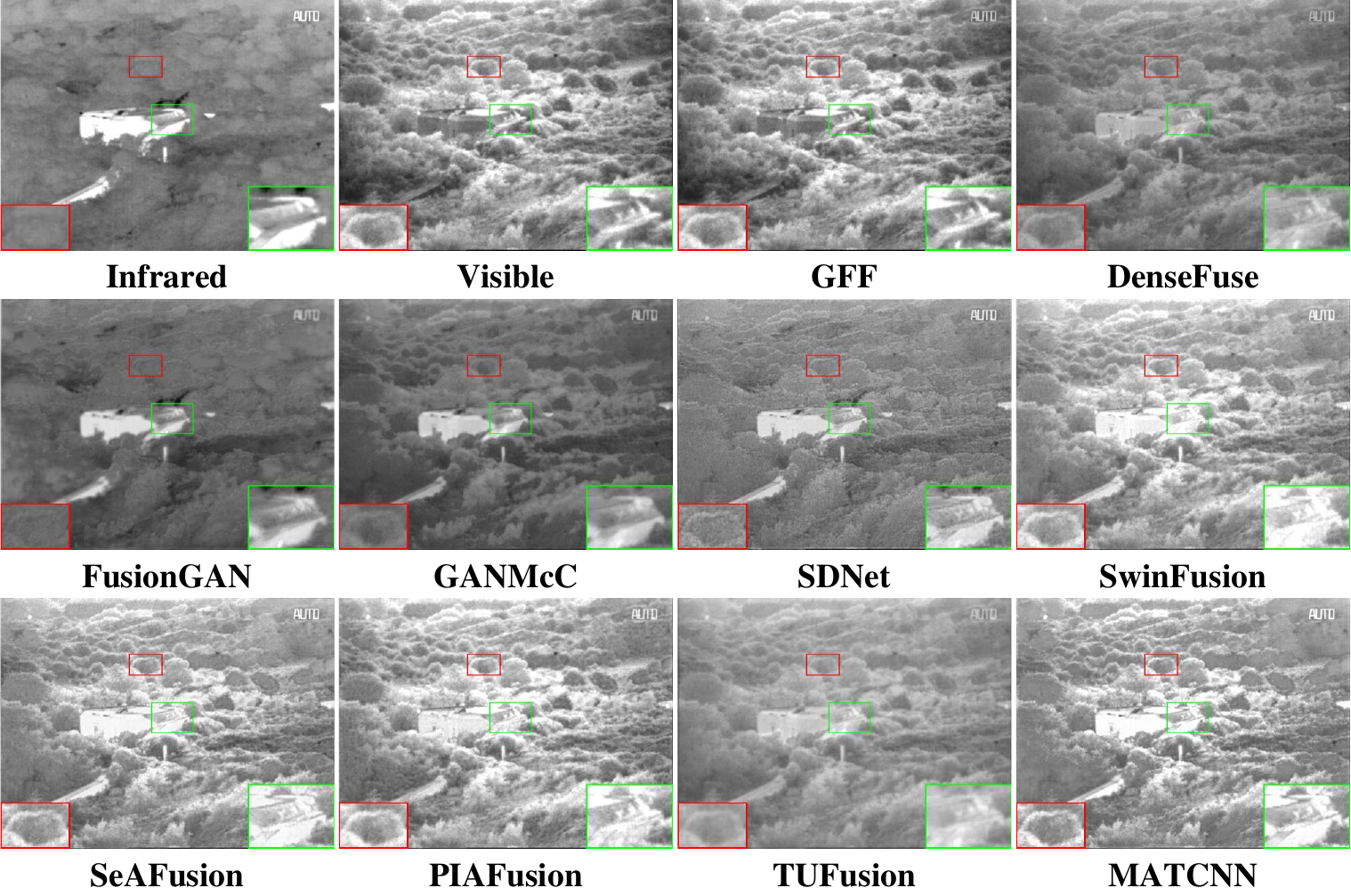}
  \caption{Fusion results of Bunker in the TNO dataset.}\label{ref_Bunker}
\end{figure}

To illustrate the variances in the fusion performance between the other methods and MATCNN intuitively, three pairs of typical images (soldiers$\_$with$\_$jeep, soldier$\_$behind$\_$smoke1, Bunker) are selected from the TNO dataset for qualitative evaluation.
The outcomes of different image fusion approaches are displayed in Fig. \ref{ref_jeep}, Fig. \ref{ref_soldier}, and Fig. \ref{ref_Bunker}.
To facilitate unambiguous comparisons, we select an area with abundant background textures using a red box, select a prominent target region using a green box, and enlarge these two areas in the bottom left and bottom right corners, respectively.

Fig. \ref{ref_jeep} shows the results of soldiers$\_$with$\_$jeep using MATCNN and other comparative fusion approaches in the TNO dataset.
In the current popular fusion methods, GFF retains background texture details but also introduces noise.
Among the current popular fusion methods, GFF, DenseFuse, SwinFusion, SeAFusion, and PIAFusion exhibit relatively smooth backgrounds, lacking texture information in the sky.
In addition, FusionGAN and GANMcC present significant distortions, resulting in blurry fusion images.
DenseFuse and TUFusion fail to sufficiently highlight infrared targets in the generated images.
Overall, MATCNN demonstrates higher contrast and clarity compared to other progressive methods, emphasizing prominent targets while better preserving texture details in visible image.

Fig. \ref{ref_soldier} shows the results of soldier$\_$behind$\_$smoke1 using MATCNN and other comparative fusion methods in the TNO dataset.
Among all the methods used for comparison, SwinFusion, SeAFusion, and PIAFusion fail to distinctly identify salient objects behind smoke.
Moreover, while GFF can identify salient objects, the contrast is low, and the salient objects lack clear prominence.
The images generated by FusionGAN and GANMcC are blurry or unclear.
TUFusion and SDNet exhibit shortcomings in preserving texture details of trees, with relatively smooth background textures.
MATCNN can maintain high contrast, retain texture details, and emphasize salient objects better than other image fusion methods, yielding superior visual effects.

Fig. \ref{ref_Bunker} shows the results of Kaptein$\_$1123 using MATCNN and other comparative fusion methods in the TNO dataset. GFF can not retain any infrared intensity features. DenseFuse and TUFusion cannot effectively highlight salient targets, images generated by FusionGAN and GANMcC are blurry or unclear, SDNet cannot preserve texture details well, and SwinFusion, SeAFusion, and PIAFusion all have a certain degree of high brightness, which cannot distinguish prominent targets from the background well. Compared to other methods, MATCNN preserves significant information and texture details while maintaining reasonable brightness, and can effectively distinguish between significant targets and background textures.

\subsubsection{Quantitative evaluation}
\begin{figure}
  \centering
  \includegraphics[width=3.5in]{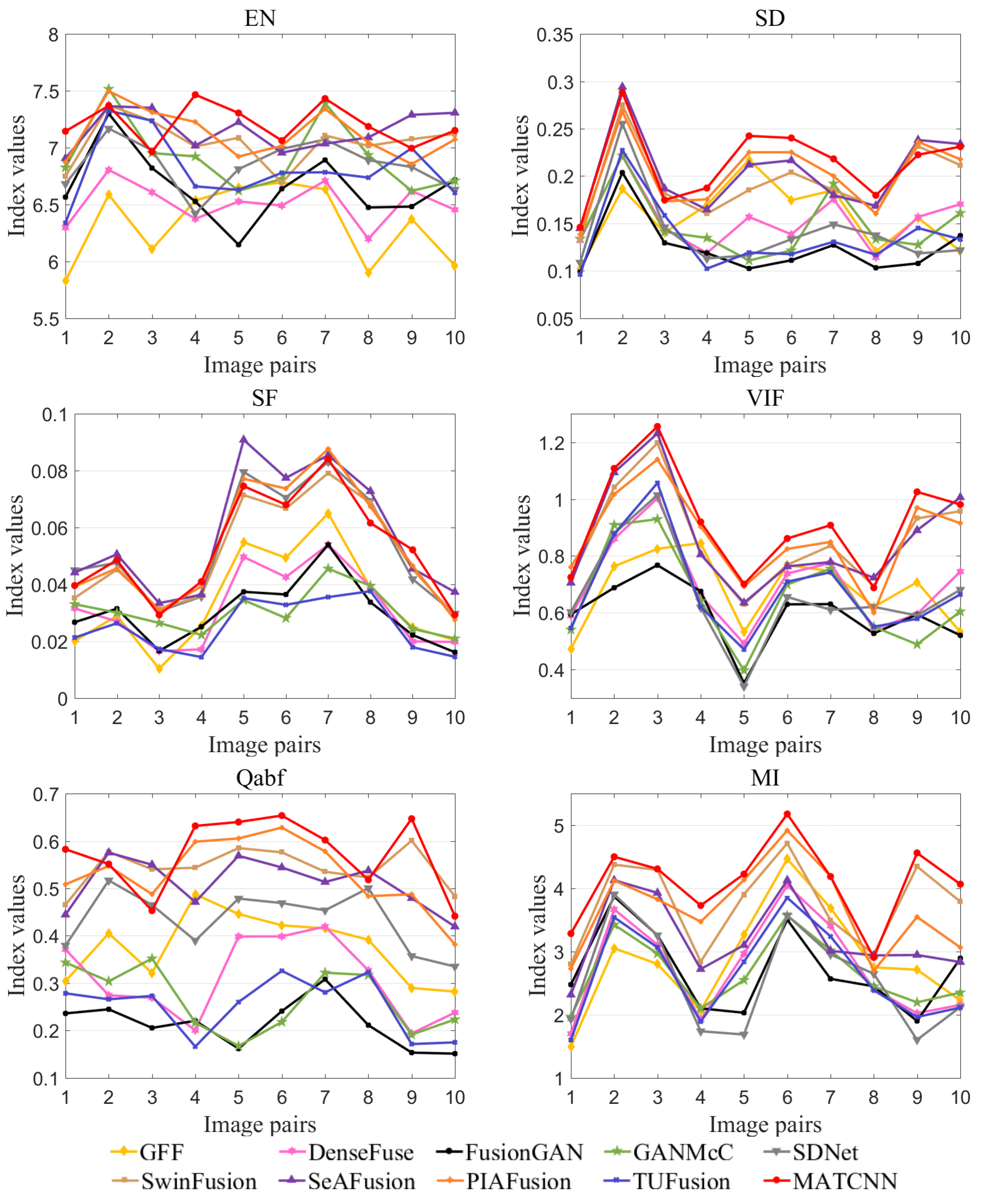}
  \caption{Objective evaluation results of the six indexes of each method in the TNO dataset.}\label{ref_index_tno}
\end{figure}

To quantitatively and objectively evaluate the image fusion performance of MATCNN in six indicators compare to the other image fusion methods, original images were randomly selected from the TNO dataset for testing.
Five random tests were conducted in total, followed by the measurement of quantitative evaluation metrics, and the average of the five tests was calculated.
For each test, utilize the system random sampling method to select 10 pairs of images.
The Fig. \ref{ref_index_tno} shows the quantitative indicator results of one of the tests, while the Table \ref{tab_avg_tno} shows the average quantitative indicator results across five tests.

As shown in the Fig. \ref{ref_index_tno}, MATCNN performs better in most indicators, especially maintaining significant advantages in MI and $Q^{AB/F}$, indicating that MATCNN can extract more valuable information from the original images and transfer the retained valuable features to the fused image.
Besides, MATCNN achieved high scores in SD and VIF, suggesting that in contrast with other image fusion approaches, the fused images produced by MATCNN have superior visual effects and higher contrast, consistent with qualitative evaluation.
Furthermore, the fused images generated by MATCNN have higher EN values compared to other image fusion methods, indicating that MATCNN is capable of generating fused images that include more information.
Although the SF value of MATCNN is not the best, it is found through comparison that MATCNN still retains sufficient edge and texture details.

To intuitively demonstrate the performance of MATCNN on each index, Table \ref{tab_avg_tno} shows the average values of the quantitative assessment indicators for the five random tests utilizing MATCNN and other approaches.
As shown in the Table \ref{tab_avg_tno}, the optimal value of every indicator is represented in bold, and the suboptimal value is indicated by an underlined.
In addition, the improvement of MATCNN compared to the optimal comparison method is calculated and shown in bracket.
MATCNN achieves the highest average score among the four indexes, including SD, VIF, $Q^{AB/F}$, and MI, only lagging behind the most advanced method in terms of EN and SF index.
The maximum increase is MI for 3.3\%, and the minimum increase is VIF for 1.2\%.
In addition, the value of SD and $Q^{AB/F}$ increased by 2.3\% and 2.8\%, respectively.
This suggests that, compared to other approaches, MATCNN can extract visible texture details and richer infrared salient targets from the original image in the TNO dataset, preserve more original information, transfer them to the fused image, and produce fused images with superior contrast and visual effects.

\begin{table*}
  \centering
  \caption{The average of six indicators in the random sampling experiment on the TNO dataset.}\label{tab_avg_tno}
  \setlength{\tabcolsep}{12 pt}
    \begin{tabular}{ccccccc}
      \toprule
      \toprule
      Methods    & EN & SD & SF & VIF & Qabf & MI \\
      \midrule
      GFF        & 6.3036 & 0.1424 & 0.04698 & 0.7201 & 0.3851 & 2.5188 \\
      DenseFuse  & 6.4237 & 0.1431 & 0.03013 & 0.7629 & 0.3123 & 2.4825 \\
      FusionGAN  & 6.5702 & 0.1371 & 0.02831 & 0.6494 & 0.2229 & 2.5091 \\
      GANMcC     & 6.7564 & 0.1340 & 0.02780 & 0.6971 & 0.2516 & 2.4487 \\
      SDNet      & 6.6857 & 0.1272 & \underline{0.05154} & 0.7191 & 0.4064 & 2.3044 \\
      SwinFusion & 6.8385 & 0.1664 & 0.04858 & 0.8594 & \underline{0.5143} & \underline{3.2899} \\
      SeAFusion  & \textbf{7.0781} & \underline{0.1870} & \textbf{0.05314} & \underline{0.8985} & 0.4711 & 2.9621 \\
      PIAFusion  & 6.8752 & 0.1744 & 0.05061 & 0.8777 & 0.5062 & 3.1991 \\
      TUFusion   & 6.5051 & 0.1143 & 0.0218 & 0.6651 & 0.2515 & 2.3796 \\
      MATCNN   & \underline{6.9862} (-1.3\%) & \textbf{0.1913} (2.3\%) & 0.05015 (-5.6\%) & \textbf{0.9094} (1.2\%) & \textbf{0.5291} (2.8\%) & \textbf{3.3978} ((3.3\%) \\
      \bottomrule
      \bottomrule
    \end{tabular}
\end{table*}

\subsection{Experiments on the MSRS dataset}
\subsubsection{Qualitative evaluation}
\begin{figure}
  \centering
  \includegraphics[width=3.5in]{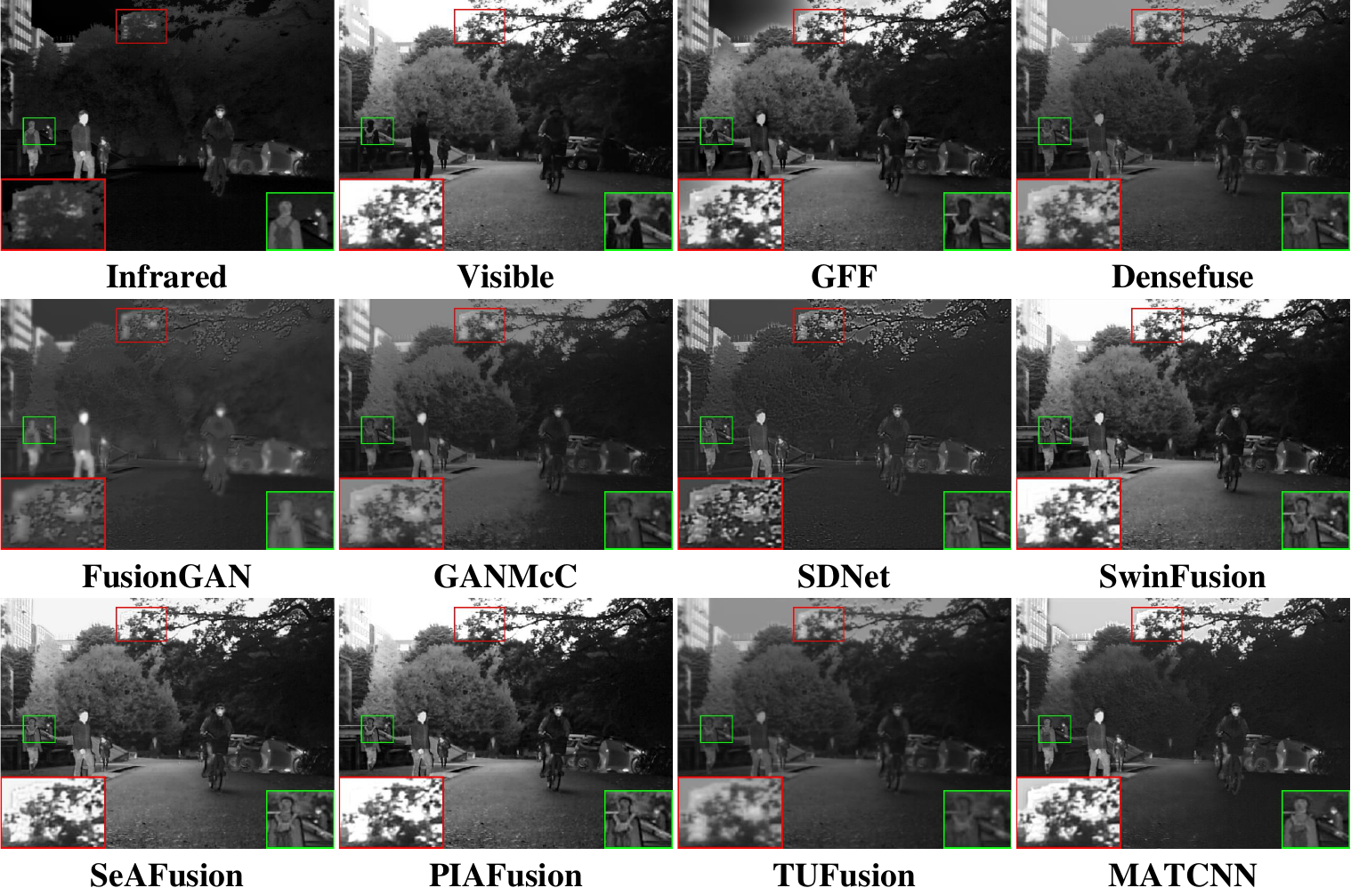}
  \caption{Fusion results of 01502D in the MSRS dataset.}\label{ref_01502D}
\end{figure}

\begin{figure}
  \centering
  \includegraphics[width=3.5in]{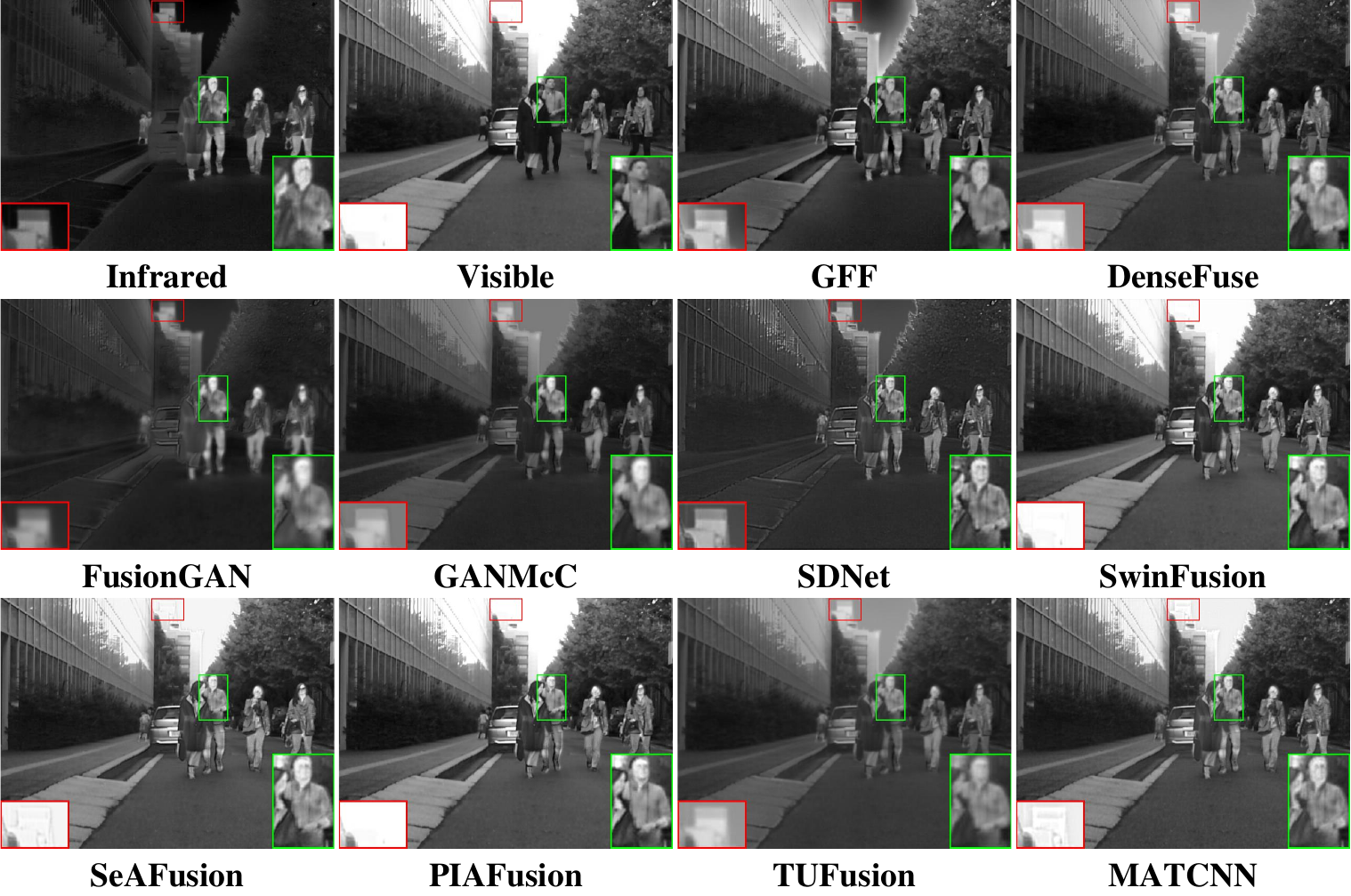}
  \caption{Fusion results of 00345D in the MSRS dataset.}\label{ref_00345D}
\end{figure}

\begin{figure}
  \centering
  \includegraphics[width=3.5in]{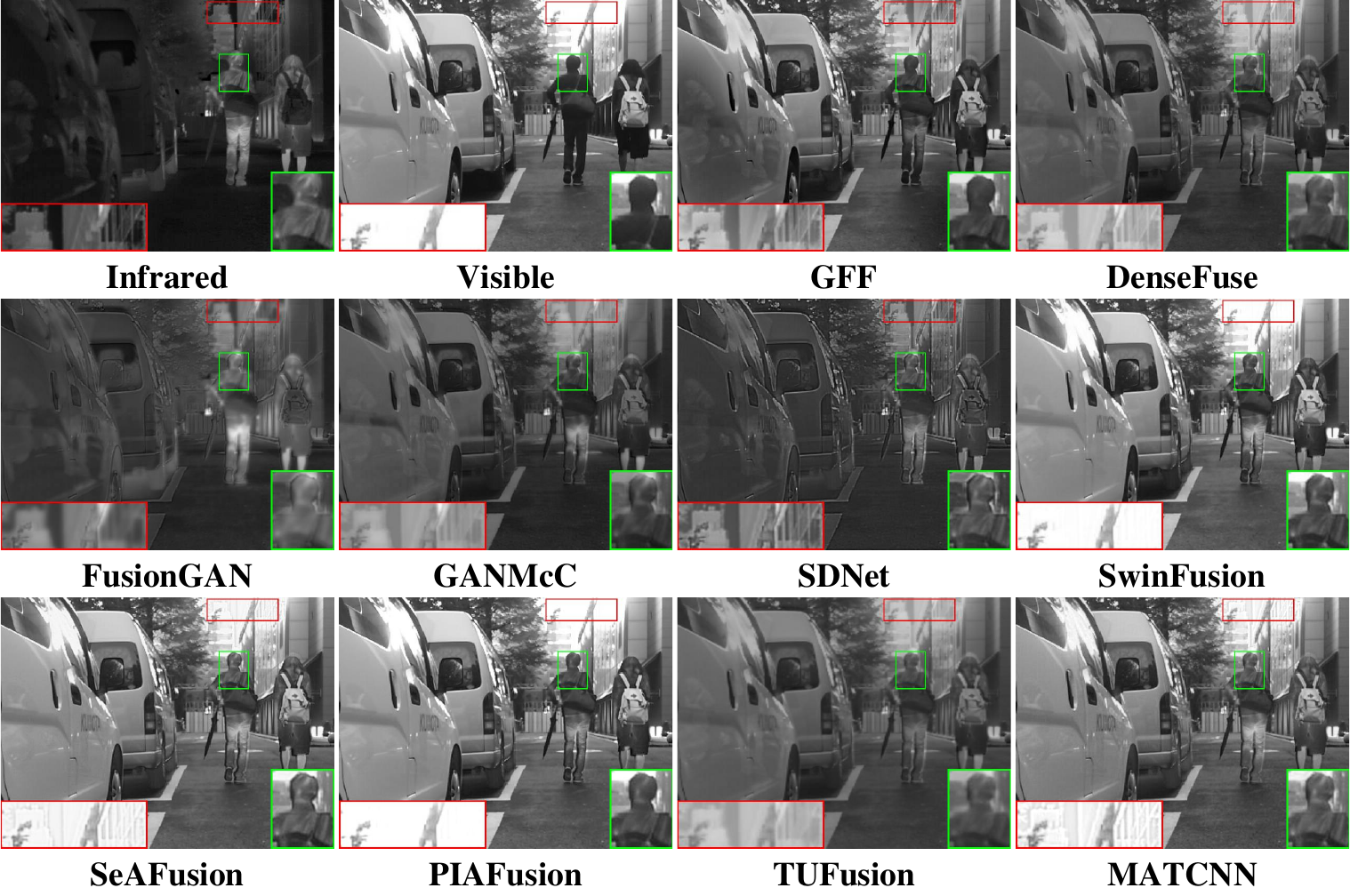}
  \caption{Fusion results of 00186D in the MSRS dataset.}\label{ref_00186D}
\end{figure}

The MSRS dataset is utilized to directly test the trained model with the objective illustrate MATCNN's ability for generalization.
Three exemplary pairings of cross-modal images (Image01502D, Image00345D, Image00186D) are selected from the MSRS dataset to intuitively notice the distinctions in the fusion performance of various approaches.
The fused results of various approaches are displayed in Fig. \ref{ref_01502D}, Fig. \ref{ref_00345D}, and Fig. \ref{ref_00186D}.

Fig. \ref{ref_01502D}, Fig. \ref{ref_00345D}, and Fig. \ref{ref_00186D} show the fusion results of Image01502D, Image00103D, Image00186D in the MSRS dataset. Among all the selected advanced algorithms, FusionGAN, GANMcC, and TUFusion have deficiencies in preserving infrared salient information, resulting in blurred salient information in the fused image. The fused images generated by SDNet existed insufficient contrast, and not in line with human visual habits.
Overall, MATCNN has advantages in preserving infrared salient information, as it can better highlight salient targets while also retaining sufficient background texture details, such as the texture of trees and buildings.

\subsubsection{Quantitative evaluation}
\begin{figure}
  \centering
  \includegraphics[width=3.5in]{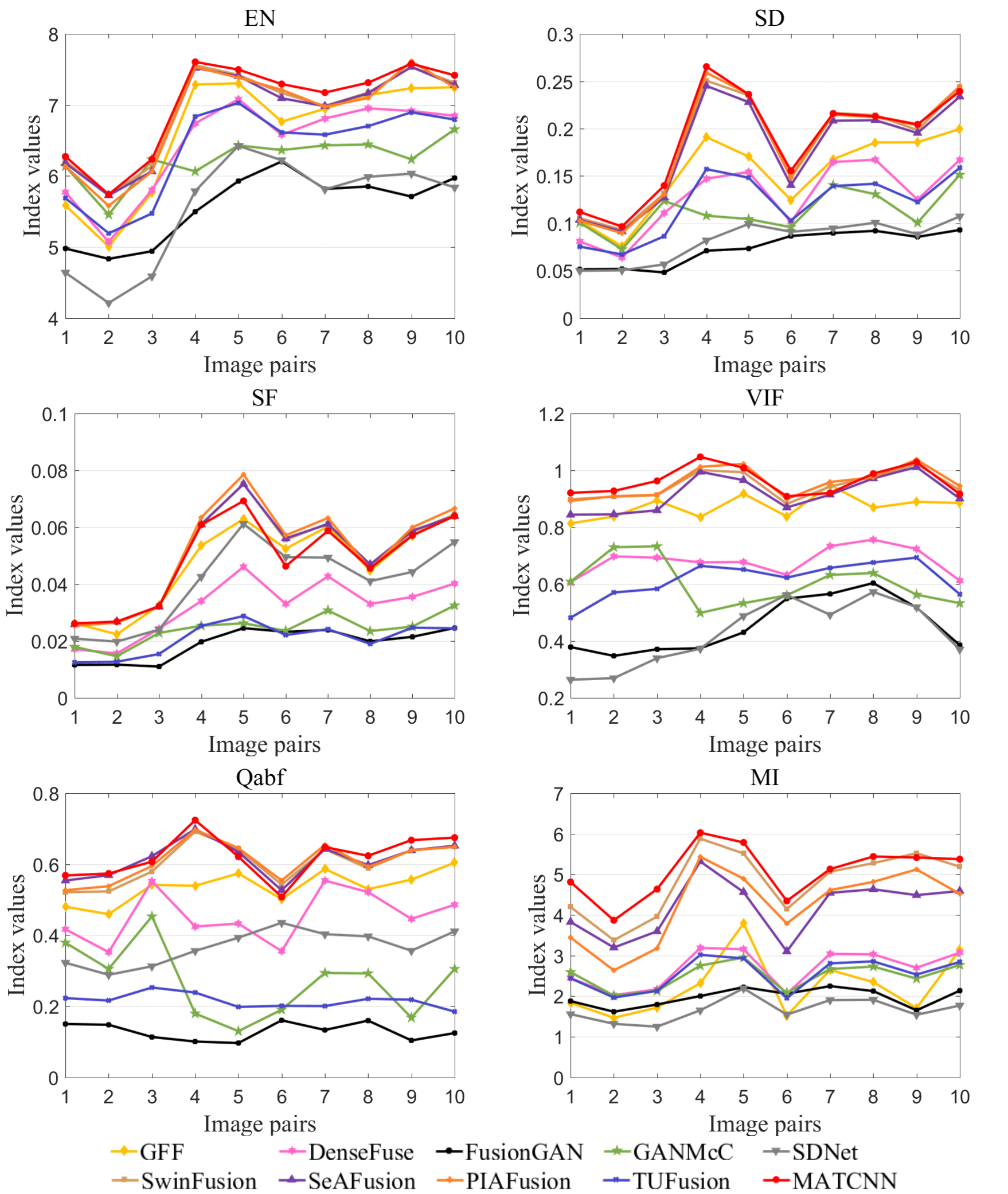}
  \caption{Objective evaluation results of the six indexes of each method in the MSRS dataset.}\label{ref_index_MSRS}
\end{figure}

To quantitatively and objectively demonstrate the image fusion outcomes of MATCNN on the MSRS dataset, five random tests were conducted, and the evaluation index results and average index are shown in the Fig. \ref{ref_index_MSRS} and Table \ref{tab_avg_MSRS}.

Fig. \ref{ref_index_MSRS} shows objective evaluation outcomes of the six indexes for every approach in MSRS dataset.
Similar to the results obtained from testing on the TNO dataset, MATCNN maintains good performance on the MSRS dataset, with advantages in most of the six indicators.
This indicates that MATCNN has good generalization ability and can effectively achieve cross-modal image fusion.

Table \ref{tab_avg_MSRS} summarizes the average values of five random tests conducted on the MSRS dataset, showcasing the average performance of the MATCNN model across various indexes.
Notably, MATCNN earns the highest average score across most indexes, except for the EN and SF index, as shown in Table \ref{tab_avg_MSRS}.
In the MSRS dataset, the MI improvement reached 3.2\%, indicating that the fused image generated by MATCNN preserves various information of the original image better than any other comparison method.
The improvement in MI reached 2.5\%, indicating that the images generated by MATCNN own better contrast compared to other approaches.
In addition, VIF and $Q^{AB/F}$ have all achieved varying degrees of improvement.
Overall, MATCNN maintains advantages in all indicators except for EN and SF, indicating that it can achieve better fusion results on different infrared and visible dataset.

\begin{table*}
  \centering
  \caption{The average of six indicators in the random sampling experiment on the MSRS dataset.}\label{tab_avg_MSRS}
  \setlength{\tabcolsep}{12 pt}
    \begin{tabular}{ccccccc}
      \toprule
      \toprule
      Methods    & EN & SD & SF & VIF & Qabf & MI \\
      \midrule
      GFF        & 6.5833 & 0.1572 & 0.04699 & 0.9079 & 0.6511 & 2.4794 \\
      DenseFuse  & 6.3611 & 0.1256 & 0.03111 & 0.7162 & 0.4544 & 2.6434 \\
      FusionGAN  & 5.6016 & 0.0789 & 0.01925 & 0.4762 & 0.1426 & 1.9511 \\
      GANMcC     & 6.1928 & 0.1115 & 0.02384 & 0.6285 & 0.2880 & 2.4758 \\
      SDNet      & 5.5019 & 0.0846 & 0.03941 & 0.4399 & 0.3748 & 1.7010 \\
      SwinFusion & \textbf{6.8644} & 0.1813 & 0.04973 & 0.9562 & 0.5701 & \underline{4.6366} \\
      SeAFusion  & 6.6367 & 0.1756 & \underline{0.04991} & 0.9395 & \underline{0.5912} & 4.0844 \\
      PIAFusion  & 6.6248 & \underline{0.1857} & \textbf{0.05062} & \underline{0.9597} & 0.5830 & 4.1054 \\
      TUFusion   & 6.4944 & 0.1587 & 0.02443 & 0.5643 & 0.1852 & 2.8444 \\
      MATCNN   & \underline{6.7987} (-0.96\%) & \textbf{0.1904} (2.5\%) & 0.04815 (-4.8\%) & \textbf{0.9735} (1.4\%) & \textbf{0.5983} (1.2\%) & \textbf{4.7847} ((3.2\%) \\
      \bottomrule
      \bottomrule
    \end{tabular}
\end{table*}

\subsection{Experiments on the RoadScene dataset}
\subsubsection{Qualitative evaluation}

\begin{figure}
  \centering
  \includegraphics[width=3.5in]{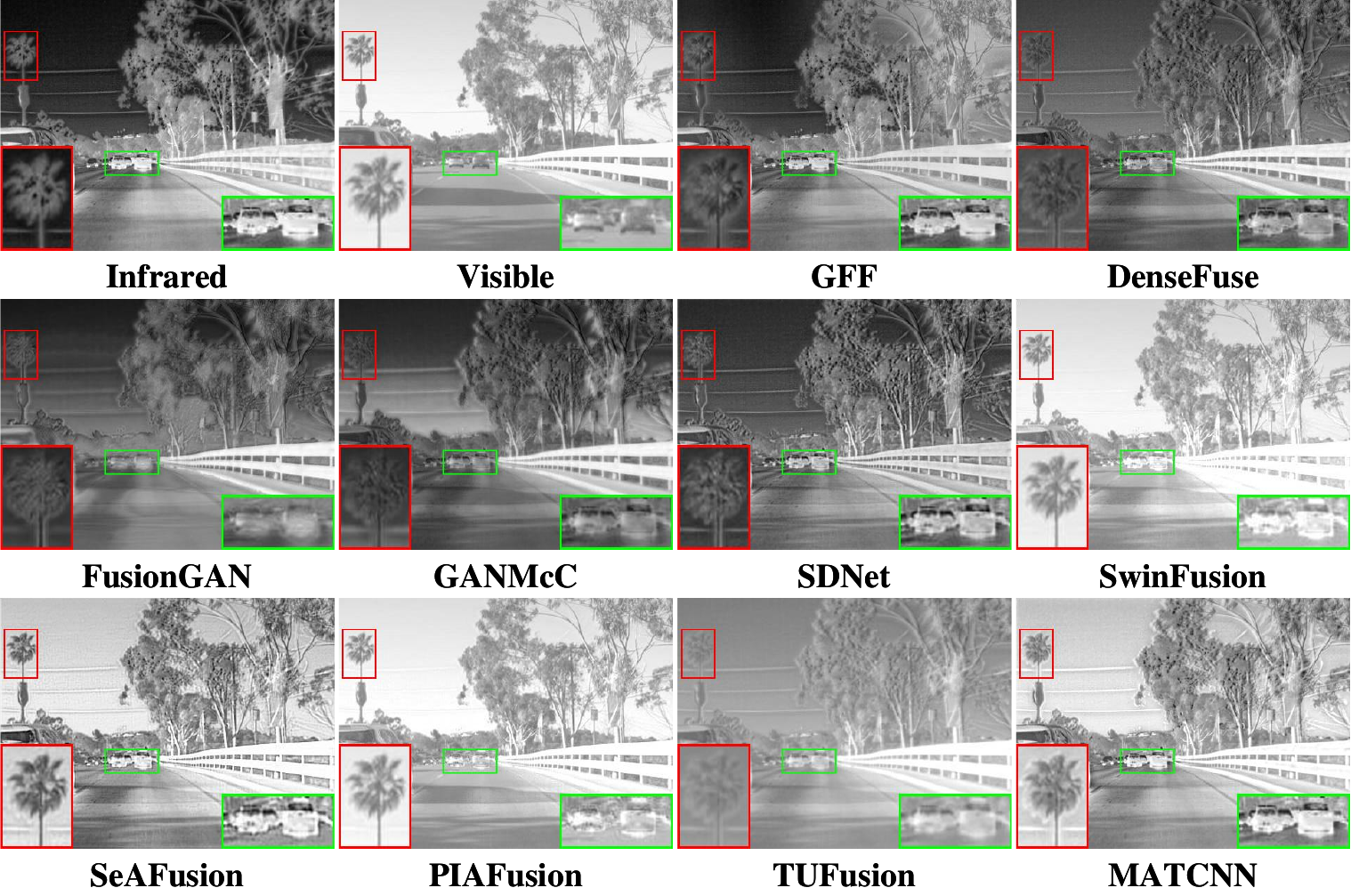}
  \caption{Fusion results of FLIR$\_$00018 in the RoadScene dataset.}\label{ref_00018}
\end{figure}

\begin{figure}
  \centering
  \includegraphics[width=3.5in]{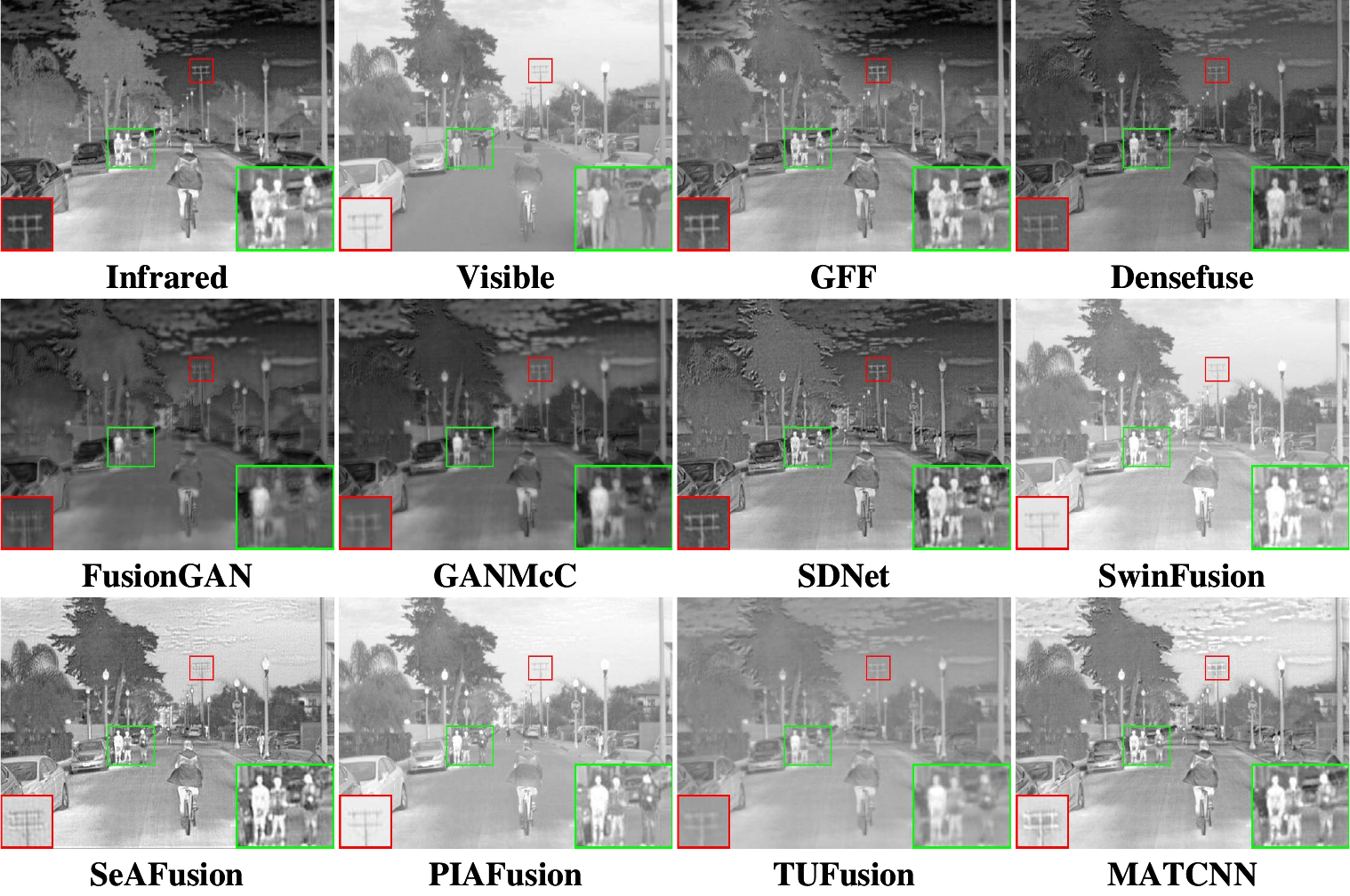}
  \caption{Fusion results of FLIR$\_$06832 in the RoadScene dataset.}\label{ref_06832}
\end{figure}

\begin{figure}
  \centering
  \includegraphics[width=3.5in]{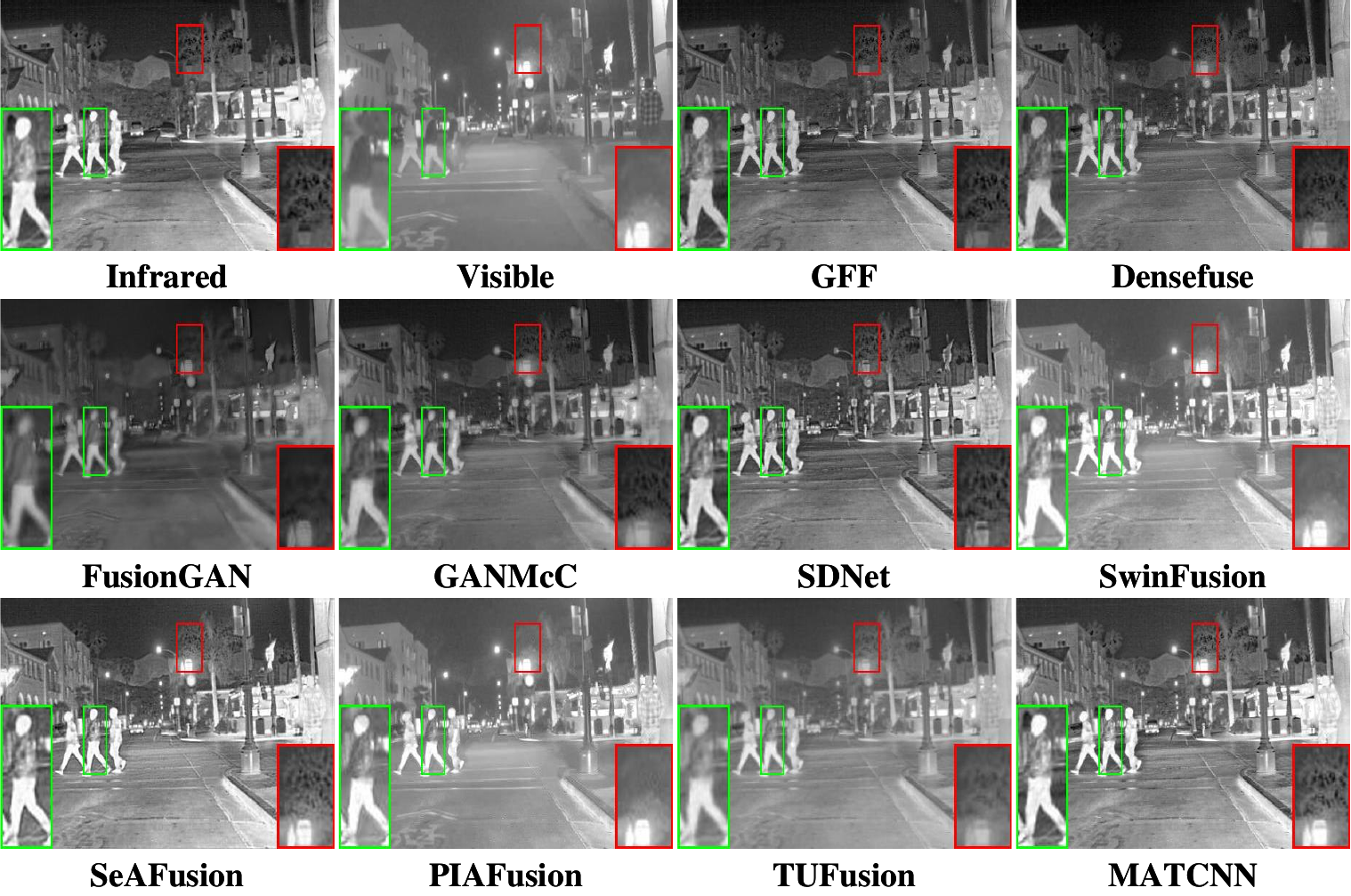}
  \caption{Fusion results of FLIR$\_$08874 in the RoadScene dataset.}\label{ref_08874}
\end{figure}

To illustrate the generalization ability of MATCNN, the trained model is directly tested RoadScene dataset without retraining.
Three pairs of typical cross-modal images (FLIR$\_$00018, FLIR$\_$06832, FLIR$\_$08874) are stochastically selected from the RoadScene dataset for qualitative evaluation.
Fig. \ref{ref_00018}, Fig. \ref{ref_06832}, and Fig. \ref{ref_08874} show the fusion outcomes of FLIR$\_$00018, FLIR$\_$06832, and FLIR$\_$08874 utilized MATCNN and other approaches.

As shown in Fig. \ref{ref_00018}, Fig. \ref{ref_06832}, and Fig. \ref{ref_08874}, among all the methods used for comparison, the sky in the fused images generated by GFF, DenseFuse, FusionGAN, GANMcC, SDNet, and TUFusion appears black or gray to varying degrees, which is not in line with human visual habits.
Furthermore, the fusion images produced by GANMcC and FusionGAN not only fail to effectively highlight prominent targets in infrared images, but also have blurry background textures, resulting in fuzzy fusion images.
The fusion images generated by SwinFusion have shortcomings in preserving background texture details in visible images, especially texture details such as trees and guardrails.
Furthermore, there is also an issue of insufficient contrast in SwinFusion.
The fusion image generated by SeaFusion can effectively highlight prominent infrared targets, have better contrast, and retain sufficient texture details.
However, the texture in the generated image is distorted, such as excessive texture in the sky, which is not correspond with human visual habits.
On the whole, MATCNN demonstrated better performance on the RoadScene dataset, achieving a balance between preserving infrared salient information and texture details, while also complying with human visual habits.

\subsubsection{Quantitative evaluation}
\begin{figure}
  \centering
  \includegraphics[width=3.5in]{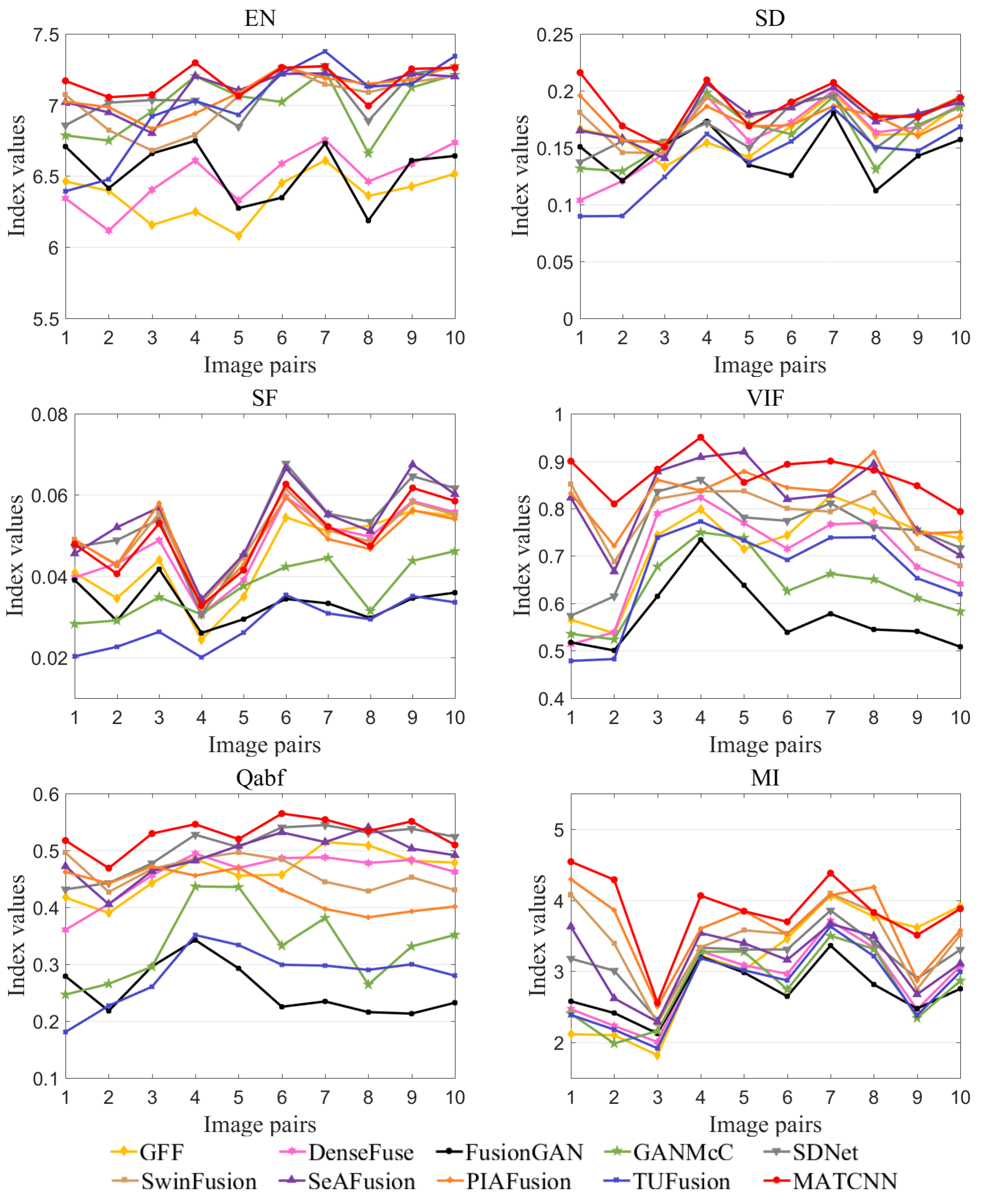}
  \caption{Objective evaluation results of the six indexes of each method in RoadScene dataset.}\label{ref_index_RoadScene}
\end{figure}

Five random sampling quantitative tests are conducted, with 10 pairs of images randomly selected from the RoadScene dataset each time using system generated random numbers.
The Fig. \ref{ref_index_RoadScene} and Table \ref{tab_avg_RoadScene} show the test results.

Fig. \ref{ref_index_RoadScene} shows the values of each indicators tested on 10 pairs of images utilized MATCNN and other approaches.
Similar to the evaluation results of testing on the tno dataset, MATCNN maintains advantage in vast majority indicators, including EN, SD, VIF, $Q^{AB/F}$ and MI, only lagging behind the most advanced algorithms in the SF indicator.
This suggests that MATCNN is more capable of generalization and can effectively fuse cross-modal images.

Table. \ref{tab_avg_RoadScene} displays the average scores of each evaluation metric for five tests on MATCNN and other methods.
As shown in the Table. \ref{tab_avg_RoadScene}, MATCNN achieves the greatest improvement in MI index, with an increase of 3.4\% compared to the most advanced methods, which indicates that MATCNN can preserve more original information in images when contrasted with other methods.
In addition, compared to the most advanced methods, MATCNN has also achieved improvement in SD, VIF, $Q^{AB/F}$ indicators, reaching 1.2\%, 2.3\% and 3.1\% respectively.
The EN index achieves the smallest improvement, only reaching 0.21\%.

Overall, on the Roadscene dataset, MATCNN can achieve better fusion results compared to other approaches, and the resulting fusion images have better visual effects and contrast, highlighting significant infrared information.

\begin{table*}
  \centering
  \caption{The average of six indicators in the random sampling experiment on the RoadScene dataset.}\label{tab_avg_RoadScene}
  \setlength{\tabcolsep}{12 pt}
    \begin{tabular}{ccccccc}
      \toprule
      \toprule
      Methods    & EN & SD & SF & VIF & Qabf & MI \\
      \midrule
      GFF        & 6.3101 & 0.1580 & 0.04813 & 0.6893 & 0.4846 & 3.0489 \\
      DenseFuse  & 6.1955 & 0.1559 & 0.04644 & 0.6771 & 0.4770 & 2.7440 \\
      FusionGAN  & 6.7653 & 0.1497 & 0.03445 & 0.5565 & 0.2725 & 2.7265 \\
      GANMcC     & 6.9276 & 0.1561 & 0.03780 & 0.6270 & 0.3449 & 2.7553 \\
      SDNet      & 6.9717 & 0.1535 & \underline{0.04937} & 0.7194 & \underline{0.5127} & 3.0994 \\
      SwinFusion & 7.0631 & 0.1620 & 0.04843 & 0.7672 & 0.4622 & 3.3126 \\
      SeAFusion  & \underline{7.0929} & 0.1613 & \textbf{0.05091} & 0.8046 & 0.5067 & 3.0464 \\
      PIAFusion  & 7.0280 & \underline{0.1635} & 0.04811 & \underline{0.8160} & 0.4361 & \underline{3.6048} \\
      TUFusion   & 6.8903 & 0.1310 & 0.02607 & 0.6328 & 0.2798 & 2.6496 \\
      MATCNN   & \textbf{7.1077} (0.21\%) & \textbf{0.1655} (1.2\%) & 0.04779 (-6.1\%) & \textbf{0.8348} (2.3\%) & \textbf{0.5284} (3.1\%) & \textbf{3.7267} ((3.4\%) \\
      \bottomrule
      \bottomrule
    \end{tabular}
\end{table*}

\subsection{Ablation experiments}
To clarify the significance of each module, ablation experiments are conducted, and the settings of control groups are shown in the Table \ref{tab_set_ablation}. Three sets of ablation experiments are conducted, namely w/o $\boldsymbol{L}_{content}$, w/o $\boldsymbol{L}_{ssim}$, and w/o $\boldsymbol{L}_{global}$.
As the roles of GFEM and Mask are to construct the loss function, we only conducted ablation experiments on the loss function.
When $\boldsymbol{L}_{global}$ is not used, GFEM is not employed.
Similarly, when $\boldsymbol{L}_{content}$ is not used, Mask is not employed.

\begin{figure}
  \centering
  \includegraphics[width=3.5in]{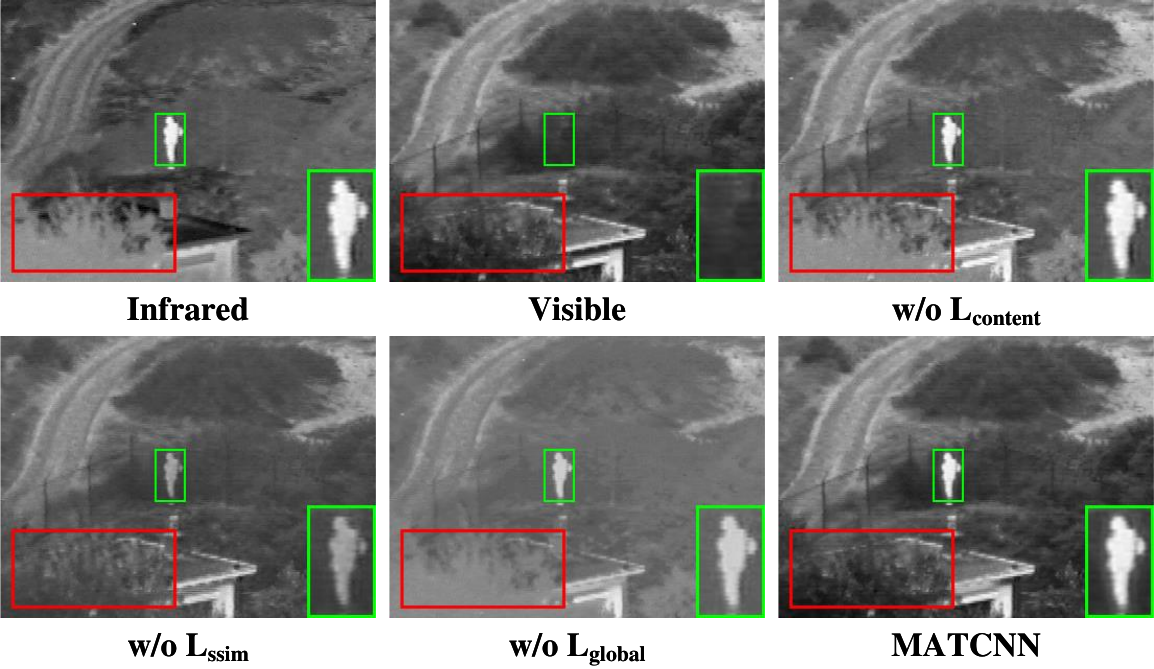}
  \caption{Fusion results of the control groups and MATCNN.}\label{ref_ablation}
\end{figure}

\begin{figure}
  \centering
  \includegraphics[width=3.5in]{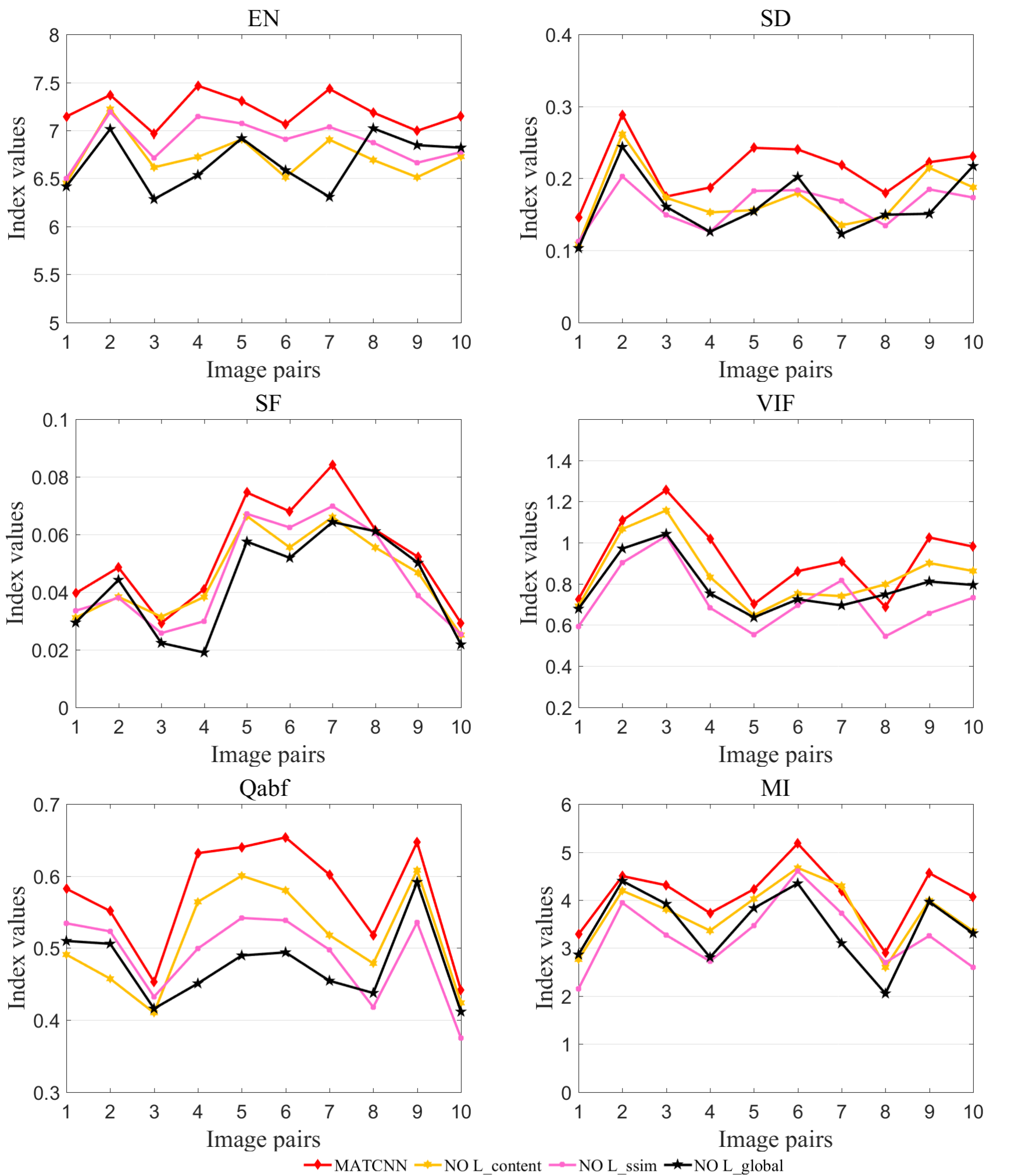}
  \caption{Objective evaluation results of each control group in the ablation experiments.}\label{ref_index_ablation}
\end{figure}

Fig. \ref{ref_ablation} shows the fusion outcomes and ablation experiments of MATCNN.
In general, the three control groups have different problems to some extent.
Because there is no annotation for useful information when $\boldsymbol{L}_{content}$ is not used, the fused images are unable to retain the texture features from the visible light images.
When $\boldsymbol{L}_{ssim}$ is not used, the fused images are lacking contrast, which makes it challenging to adequately emphasize the salient targets in the infrared images.
When $\boldsymbol{L}_{global}$ is not used, the infrared image targets exhibit noticeable blurriness, and the background texture detail is insufficient.
The objective assessment results for each control group in the ablation experiments are displayed in Fig. \ref{ref_index_ablation}, which are to some extent lower than MATCNN.
It is evident that the proposed loss function, which encompasses content loss, structure similarity loss, and content loss, can enhance the visual impact of the fusion of cross-modal images and contribute positively to the generated images.

\begin{table}
  \centering
  \caption{The Control Group Set In Ablation Experiments}\label{tab_set_ablation}
  \setlength{\tabcolsep}{12 pt}
    \begin{tabular}{cccc}
      \toprule
      \toprule
      Control Groups    & $\boldsymbol{L}_{content}$ & $\boldsymbol{L}_{ssim}$ & $\boldsymbol{L}_{global}$ \\
      \midrule
      w/o $\boldsymbol{L}_{content}$  & \XSolidBrush & \Checkmark & \Checkmark \\
      w/o $\boldsymbol{L}_{ssim}$ & \Checkmark & \XSolidBrush & \Checkmark \\
      w/o $\boldsymbol{L}_{global}$  & \Checkmark & \Checkmark & \XSolidBrush \\
      \bottomrule
      \bottomrule
    \end{tabular}
\end{table}

\section{Conclusion}\label{conclusion}

This paper presents a novel cross-modal image fusion approach that combines multi-scale CNN and attention Transformer, overcoming the shortcomings of extracting multi-scale local features and preserving global features in existing fusion methods.
MATCNN designs a novel feature extraction model that utilizes an MSFM to extract local features at various scales.
Firstly, it extracts features from original cross-modal images through the trunk, inputs them into three branches for fusion, and continues extracting features.
Finally, the feature information obtained from the trunk and branches is fused to acquire the fused image.
In addition, a GFEM is designed based on attention Transformer to extract global features from input and fused images, and a novel loss function of global is used to estimate the global features differences between input and fused images at four different scales, improving the feature continuity of fused images.
Compared with other latest approaches, the fusion image generated by MATCNN has better contrast and highlighted better prominent infrared targets, the quantitative evaluation index has increased by a maximum of 3.3\%.
In addition, visible images also contain a small amount of important information, while infrared images provide a small amount of necessary background textures.
MATCNN uses infrared salient target masks to annotate salient information in infrared images, which may affect the extraction of background textures within infrared images as well as the salient information within visible images.
In future work, we will strive to address this problem.

\bibliographystyle{IEEEtran}
\bibliography{MATCNN}

\begin{IEEEbiography}[{\includegraphics[width=1in,height=1.25in,clip,keepaspectratio]{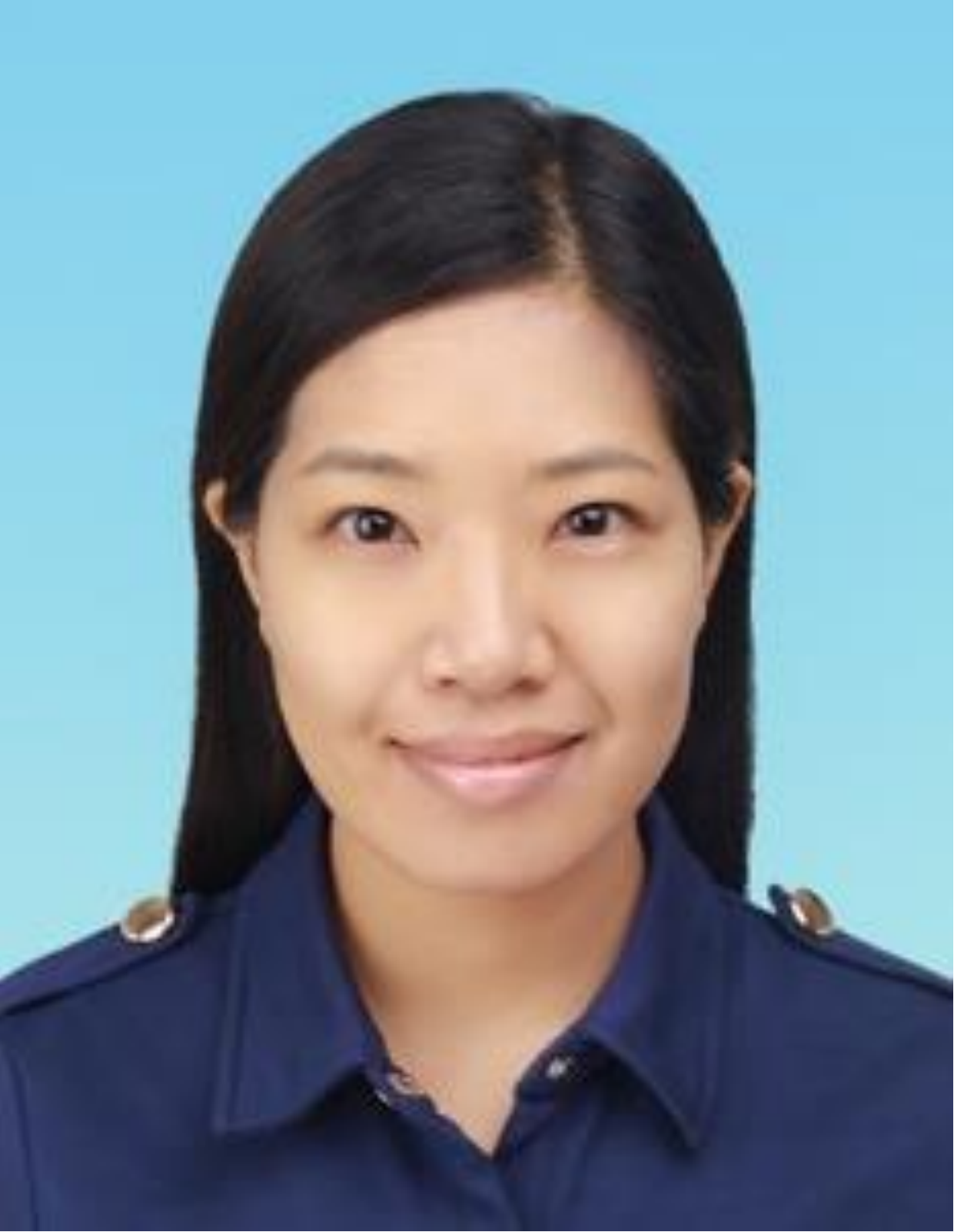}}]{Dr. Jingjing Liu}
received her Bachelor degree in Electrical Engineering from Zhengzhou University, P.R. China, in 2006, the MSc and PhD degree in Control Science and Engineering from Shanghai University, Shanghai, China, in 2013 and 2018. She was a joint PhD in Computation department of Curtin University in Australia from December 2015 to December 2016. She was a post-doctoral at the State Key Laboratory of ASIC and System, School of Microelectronics, Fudan University, from 2019 to 2022. Currently, she is a faculty member at School of Microelectronics, Shanghai University, with interests in artificial intelligence algorithms and chip design.
\end{IEEEbiography}

\begin{IEEEbiography}[{\includegraphics[width=1in,height=1.25in,clip,keepaspectratio]{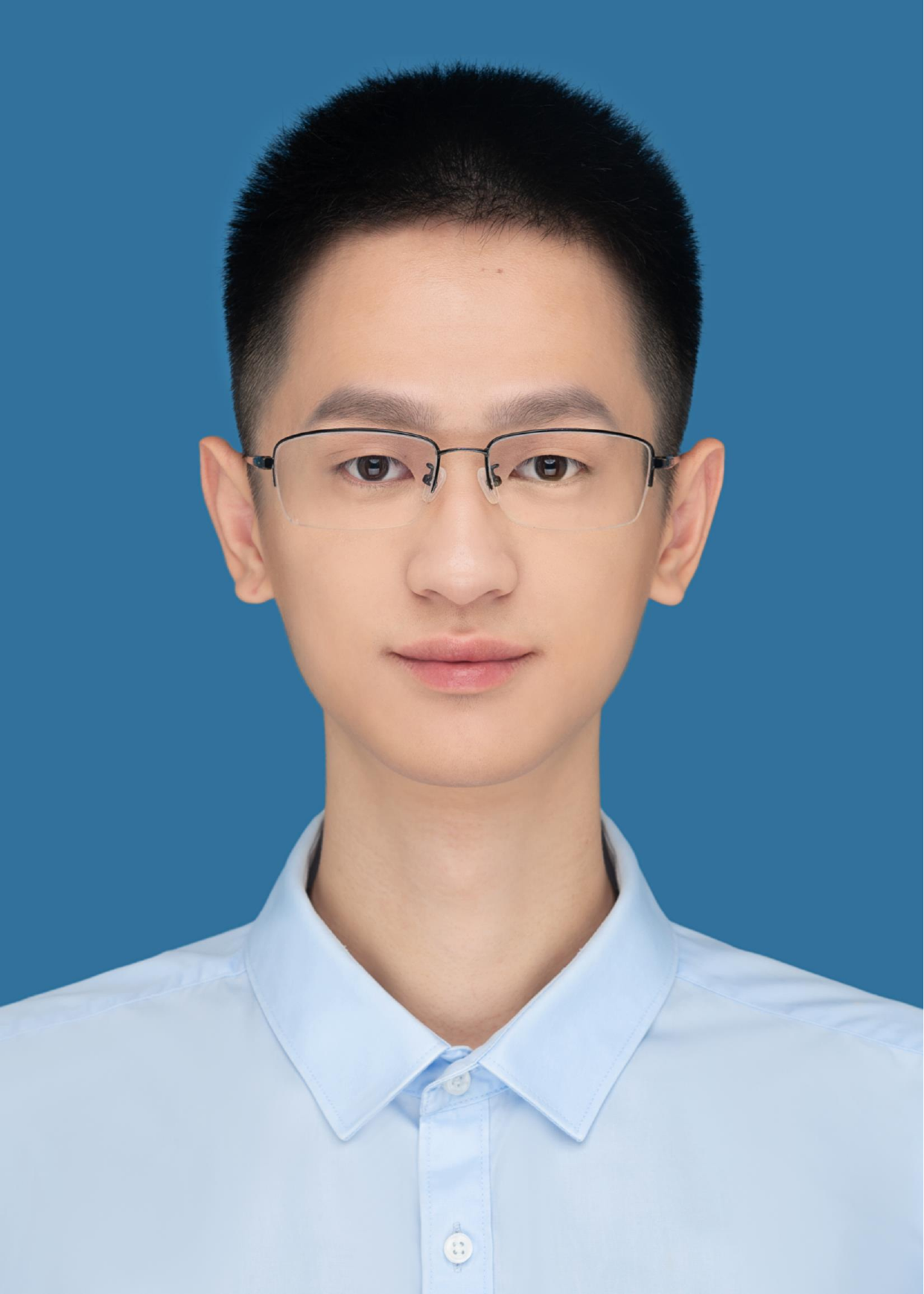}}]{Li Zhang}
received the Bachelor degree in Optoelectronic Information Science and Engineering from Harbin Institute of Technology, P.R. China,in 2020. He is currently pursuing the MSc degree in Electronic Information at the School of Microelectronics, Shanghai University. His current research interests include image fusion algorithms and hardware implementation.
\end{IEEEbiography}

\begin{IEEEbiography}[{\includegraphics[width=1in,height=1.25in,clip,keepaspectratio]{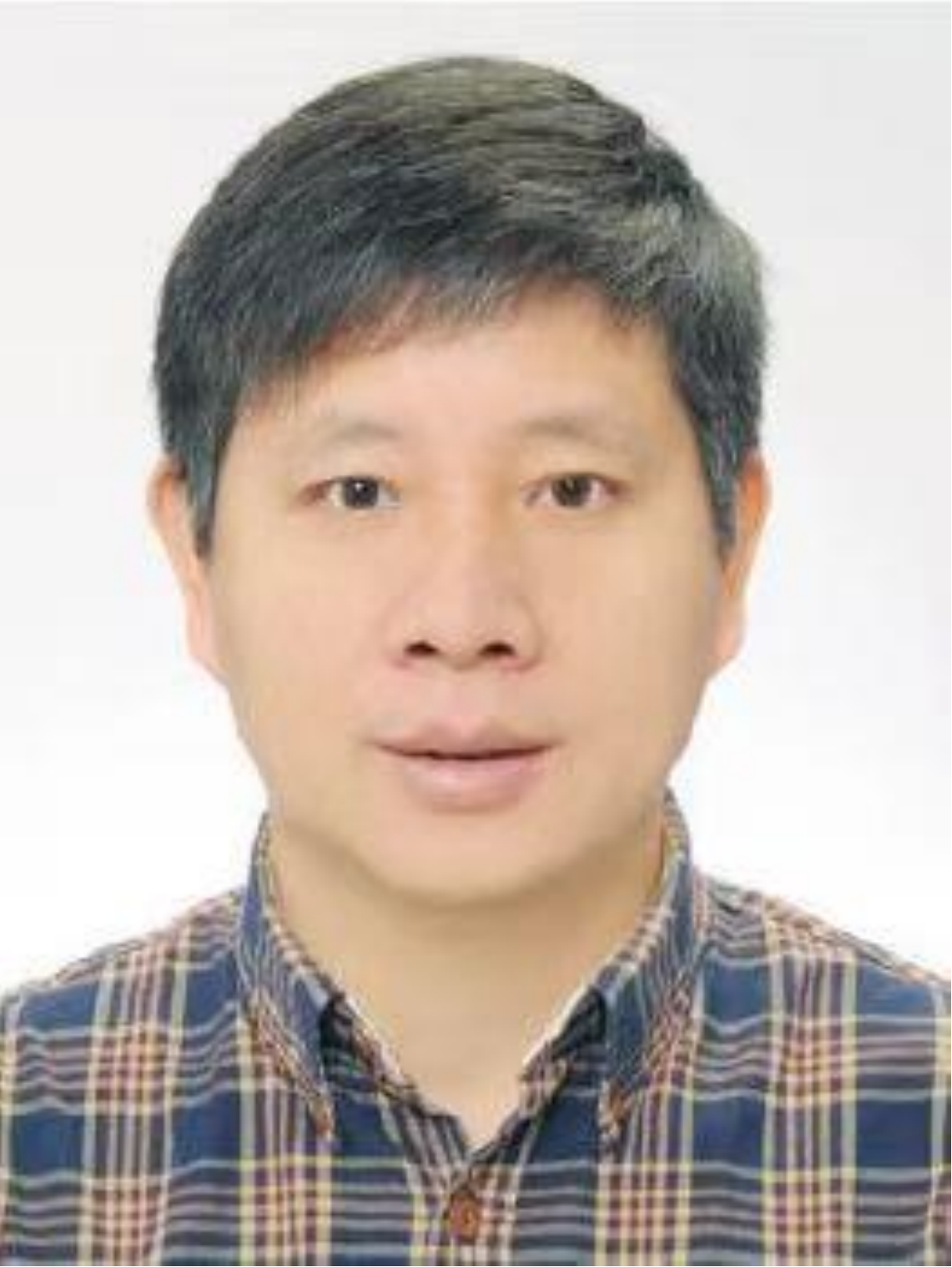}}]{Prof. Xiaoyang Zeng}
received the Ph.D. degree from the Chinese Academy of Sciences. Beijing, China, in 2001. He is currently the Executive Director of the State Key Laboratory of ASIC and System and the Vice Dean of the School of Microelectronics, Fudan University, Shanghai. His research interests include high-performance and low-power VLSl architecture design for information security algorithms, digital signal processing algorithms, wireless communication base-band processing, and the mixed-signal circuits design technology. Dr. Zeng also serves as the Co-Chair of the Circuit and System Division. Chinese Institute of Electronics; the Steering Committee Member of ASP.DAC: a TPC Member of A-SsCC: and the TPC Chair of ASICON 2009/2013.
\end{IEEEbiography}

\begin{IEEEbiography}[{\includegraphics[width=1in,height=1.25in,clip,keepaspectratio]{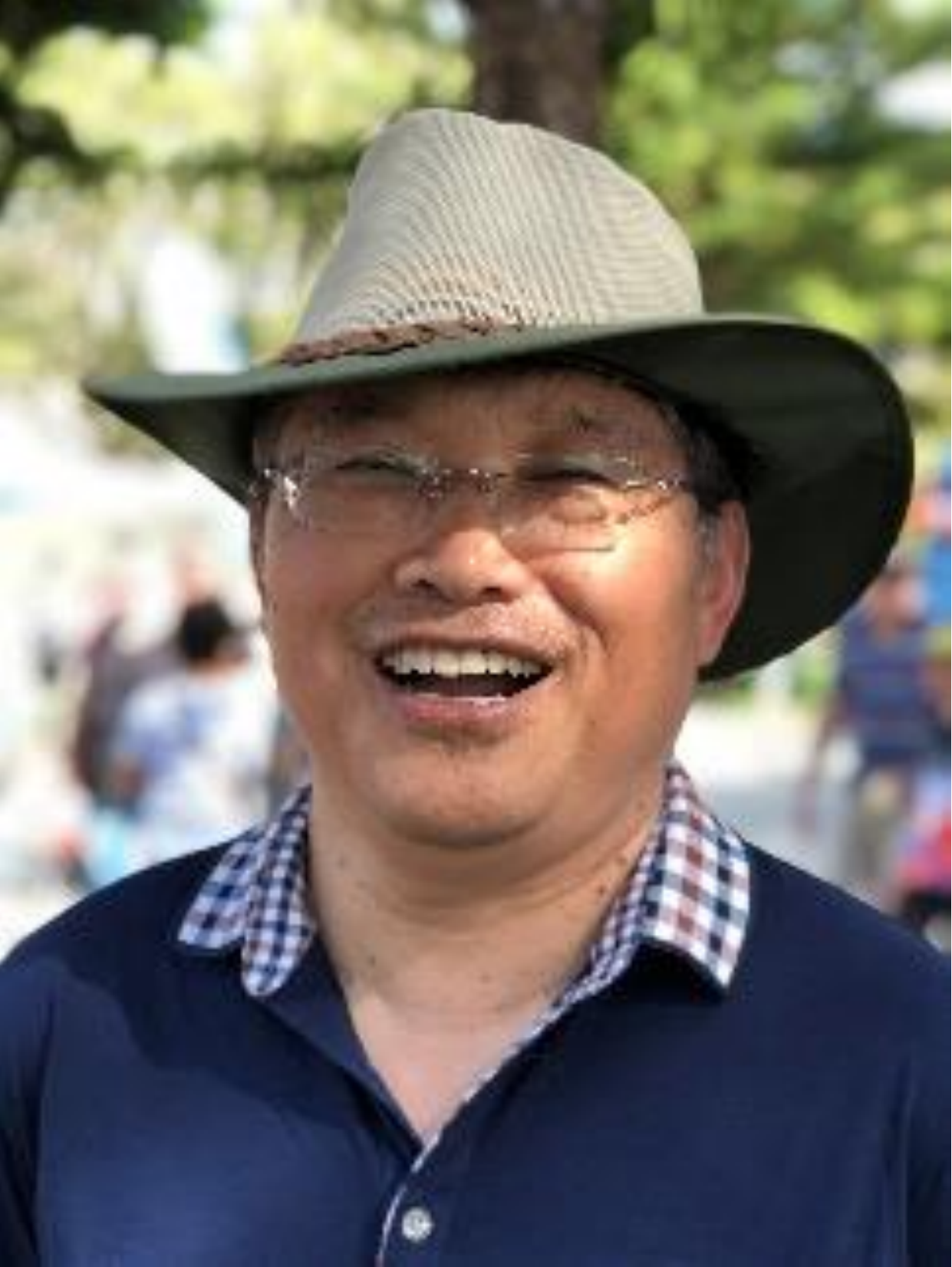}}]{Prof. Wanquan Liu}
received the PhD degree in electrical engineering from Shanghai Jiaotong University, in 1993. He once held the ARC Fellowship, U2000 Fellowship, and JSPS Fellowship and attracted research funds from different resources more than 3.5 million Australian dollars. He is currently a Professor in School of Intelligent Systems Engineering at Sun Yat-sen University. He is the Editor-in-chief for the international journal Mathematical Foundation of Computing and in editorial board for seven international journals. His current research interests include large-scale pattern recognition, signal processing, control engineering, machine learning, and computer vision.
\end{IEEEbiography}

\begin{IEEEbiography}[{\includegraphics[width=1in,height=1.25in,clip,keepaspectratio]{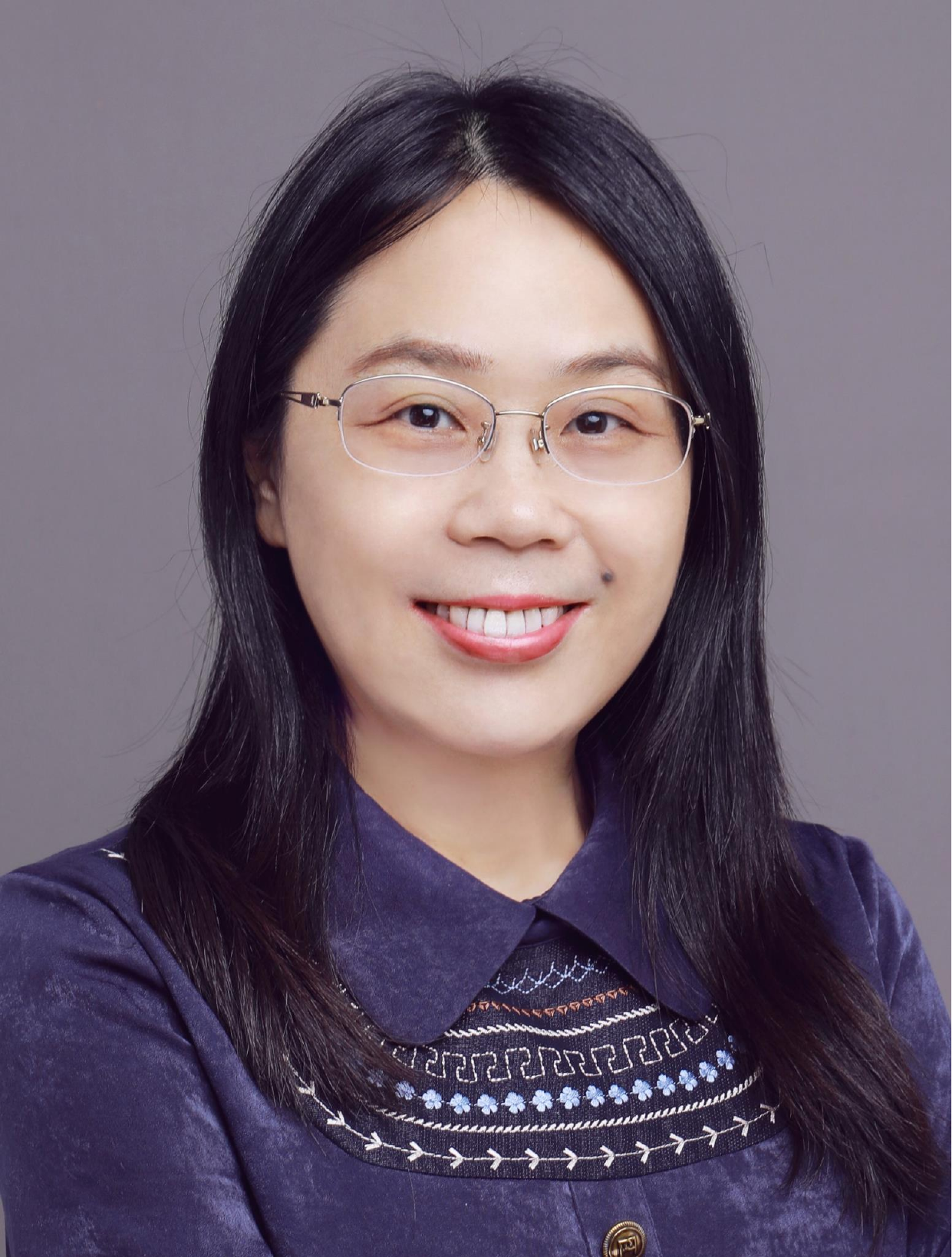}}]{Prof. Jianhua Zhang}
received the Ph.D. degree in mechanical engineering from Shanghai University, Shanghai, China, in 1999.
She is currently a Professor with the School of Microelectronics, Shanghai University, where she is also the Director of the Key Laboratory of Advanced Display and System Applications. Her main research interests include digital chip design and system integration, digital video and stereoscopic display technology, and artificial intelligence algorithms and chip design.
Prof. Zhang was a recipient of the First Prize of Shanghai Award for Technological Invention (2014, First order), the Prize of Shanghai Award for Scientific and Technological Progress (2016 and 2020, First order), and also the China National Funds for Distinguished Young Scientists in 2017.
\end{IEEEbiography}

\end{document}